\documentclass[lettersize,journal]{IEEEtran}
\usepackage{amsmath,amsfonts}
\usepackage{algorithmic}
\usepackage{array}
\usepackage{textcomp}
\usepackage{stfloats}
\usepackage{url}
\usepackage{verbatim}
\usepackage{graphicx}
\usepackage{cite}
\usepackage{bm}
\usepackage{subcaption}
\usepackage{hyperref}

\usepackage{tabularx}
\usepackage{algorithm}

\usepackage{pgfplots}
\pgfplotsset{compat=1.18}
\usepgfplotslibrary{groupplots}
\usepgfplotslibrary{dateplot}
\usetikzlibrary{arrows.meta}
\usetikzlibrary{calc}
\usetikzlibrary{backgrounds, plotmarks}
\usetikzlibrary{decorations.markings}
\usetikzlibrary{positioning}
\usetikzlibrary{fit}
\usetikzlibrary{patterns,patterns.meta}
\usetikzlibrary{decorations.text, chains, shapes.geometric, intersections}

\usepackage{amsthm_custom}  
\usepackage{mycustompgf}
\usepackage{mydefinitions}
\usepackage[short, c2, nocomma]{optidef_custom}

\newtheorem{proposition}{Proposition}
\newtheorem{example}{Example}

\newtheorem{remark}{Remark}
\newtheorem{assumption}{Assumption}

\begin{document}

\title{Robust Rigid Body Assembly via Contact-Implicit \\ Optimal Control with Exact Second-Order Derivatives}

\author{Christian Dietz${}^{1,2}$, Sebastian Albrecht${}^{1}$, Gianluca Frison${}^{2}$, Moritz Diehl${}^{2,3}$, Armin Nurkanovi\'c${}^{2}$
	\thanks{\noindent \hspace*{-1.06em}${}^{1}$Autonomous Systems and Control, Siemens AG, Germany \newline
		${}^{2}$Department of Microsystems Engineering (IMTEK), University of Freiburg, Germany \newline
		${}^{3}$Department of Mathematics, University of Freiburg, Germany \newline 
		Correspondent: {\tt\small dietz.christian@siemens.com}
}}%

\markboth{}%
{}


\maketitle

\begin{abstract}
Efficient planning of assembly motions is a long standing challenge in the field of robotics that has been primarily tackled with reinforcement learning and sampling-based methods by using extensive physics simulations. This paper proposes a sample-efficient robust optimal control approach for the determination of assembly motions, which requires significantly less physics simulation steps during planning through the efficient use of derivative information. To this end, a differentiable physics simulation is constructed that provides second-order analytic derivatives to the numerical solver and allows one to traverse seamlessly from informative derivatives to accurate contact simulation. The solution of the physics simulation problem is made differentiable by using smoothing inspired by interior-point methods applied to both the collision detection as well as the contact resolution problem. We propose a modified variant of an optimization-based formulation of collision detection formulated as a linear program and present an efficient implementation for the nominal evaluation and corresponding first- and second-order derivatives. Moreover, a multi-scenario-based trajectory optimization problem that ensures robustness with respect to sim-to-real mismatches is derived. The capability of the considered formulation is illustrated by results where over 99\% successful executions are achieved in real-world experiments. Thereby, we carefully investigate the effect of smooth approximations of the contact dynamics and robust modeling on the success rates. Furthermore, the method's capability is tested on different peg-in-hole problems in simulation to show the benefit of using exact Hessians over commonly used Hessian approximations.
\end{abstract}


\section{Introduction}
\IEEEPARstart{D}{ifferentiable} physics simulation increasingly receives attention in the robotics community motivated by the potential that providing derivative information to planning and control algorithms can substantially increase their computational efficiency \cite{Levine13, Xu2022,Luo2024FoPG}. It is already successfully leveraged within reinforcement learning (RL) and model predictive control (MPC) methods applied to tasks such as quadruped walking or dexterous manipulation \cite{Tassa2012, Kim2024KaistMPC}. Assembly problems have predominantly been addressed by derivative-free reinforcement learning algorithms using extensive simulation \cite{Vuong2023, Lee2024} even though derivative-based methods showed improved sample-efficiency on other problem classes \cite{Xu2022,Luo2024FoPG}. We deduce that there is currently a gap of derivative-based algorithms that successfully utilize differentiable physics simulation methods to solve assembly tasks.

\begin{figure}
	\centering
	\includegraphics[width=\linewidth]{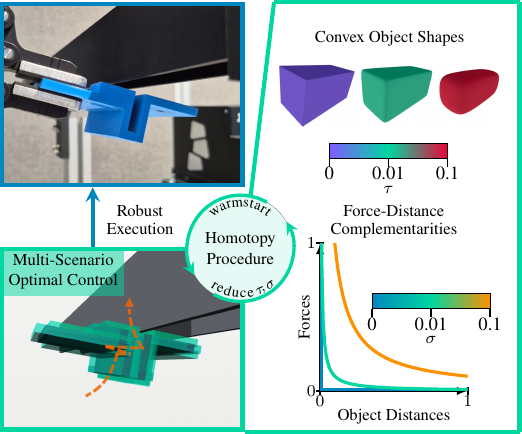}
	\caption{Illustration of a considered assembly problem and this paper's methodology. The optimization of assembly trajectories is facilitated by initially relaxing object shapes and force-distance complementarities followed by sequentially tightening the relaxations. Video highlighting the methodology: \url{https://youtu.be/g4E83bjs7lg}}
	\label{fig:title}
\end{figure}

The large majority of the work on physics simulation for robotics agrees on formulating contact-rich dynamical systems in continuous time as complementarity Lagrangian
systems with state jumps \cite{Brogliato2016}, and deriving discrete-time dynamics by applying a time-stepping discretization to obtain a nonlinear complementarity problem (NCP) \cite{Moreau}. To construct the contact NCP, first multiple collision detection problems that compute the distance between pairs of convex objects have to be solved and corresponding contact normal vectors have to be determined which encode the direction in which contact forces act. These collision detection problems for convex objects can be formulated as convex quadratic programs such that their optimality conditions take the form of linear complementarity problems (LCP). Thus, computing a single simulation step comes down to solving multiple collision LCP and then using these solutions to construct and solve a single contact NCP.

The challenge for using derivative-based methods to solve contact-rich tasks lies primarily in the fact that the solution of the contact NCP is inherently nonsmooth with respect to its problem data. This means that the position and velocity of objects at the next time step can be nondifferentiable with respect to the position and velocity of objects as well as the control input at the current time step. This is due to two hierarchically structured causes. The first cause are positions, where an infinitesimal alteration results in a change of an active complementarity condition in a collision LCP. This means contact points switching between faces, edges or vertices on a convex object. The second cause are current positions and velocities where infinitesimal alterations result in changes of active complementarity conditions in the contact NCP. Here two distinctions can be made. Either such a switch corresponds to activation or deactivation of contact forces between a pair of convex objects, or one pair of such objects switches its friction mode from static to sliding or vice versa.

Consequentially, constructing differentiable physics simulations necessarily requires additional custom design choices, as for the original NCP formulation derivatives do not exist everywhere. There are three common choices that can be found in the physics simulation literature. First, determining exact derivatives under the assumption that active complementarity conditions do not change \cite{Lelidec2025, Werling2021Nimble}. For situations where infinitesimal changes in positions or velocities result in switches in the active complementarity conditions, one obtains subgradients. Second, using finite differences or randomized smoothing \cite{Montaut2023, Lelidic2024randomizedsmoothing} to obtain approximate derivatives that do not correspond to exact derivatives of the nominal LCP or NCP evaluation. These two approaches both likely hinder success of optimization routines, as either the produced derivatives are not globally informative for the behavior of the underlying dynamics, or there is a mismatch between the nominal evaluation that is nonsmooth and the derivatives that correspond to a smooth approximation of the nonsmooth evaluation. \\
A more optimization-suited approach is therefore the third, which is constructing smooth approximations to the complementarity problems and then determining corresponding exact derivatives \cite{Howell2023Dojo, Tracy2023}. Remarkably, most existing work on differentiable physics simulation only focuses on the nondifferentiabilities caused by the contact NCP while disregarding the contribution of the collision LCP. This likely contributes to the lack of work that tackles assembly problems with complex geometries using differentiable physics simulation.

In this paper, we focus on frictionless contacts which leads to the reduction of the contact NCP to a contact LCP, under the precondition that a semi-implicit Euler integration was used to derive the time-stepping discretization. A LCP formulation of contact dynamics can also be obtained when friction is considered by linearization of the friction cone as done in several physics engines \cite{Werling2021Nimble}. We formulate the collision detection problem as a linear program (LP) using so-called growth distances \cite{Ong1996}, where we propose a modified formulation that is motivated by the goal to obtain well-scaled derivatives that can be utilized in optimal control formulations. We construct smooth approximations to this distance problem by leveraging interior-point methods. Recurrent application of the implicit function theorem enables analytic formulations for first- and second-order derivatives. \\
For the contact resolution problem, the well-established quadratic program formulation of Moreau \cite{Moreau} is used, which allows one to obtain a smooth approximation again by utilizing smoothing as implied by interior-point methods. We take the approach of directly transcribing the optimality conditions of the contact resolution problem into the optimal control problem. In this case, the interior-point smoothing can also be interpreted as Scholtes' smoothing applied to the complementarity conditions of the contact LCP and also enables one to use Scholtes' relaxation method \cite{Scholtes} which relaxes each complementarity condition individually. These two alternatives are compared within numerical experiments, emphasizing that control algorithms using differentiable physics simulation can profit from additional degrees of freedom that enables the optimization solver to control the physical evolution.\\
To ensure that the computed assembly motions can be reliably executed on real robotic systems, we utilize a multi-scenario-based optimal control formulation with a state-feedback control law given by a Cartesian impedance controller. We consider assembly problems with almost zero clearance. On the one hand, one then needs to obtain solutions for simulations where the collision and contact LCP are solved to high accuracy to ensure a reliable transfer to the real-world. On the other hand, substantial smoothing of these LCP results in nonstiff contact dynamics, enabling optimization solver convergence from trivial initial guesses. To get the best out of two worlds, we use a homotopy procedure that sequentially tightens the relaxations and uses previous solutions to warmstart the optimization solver, enabling the computation of solutions for very stiff contact dynamics which closely approximate the original nondifferentiable formulations. \\
We carefully investigate the effect that the amount of contact dynamics smoothing as well as robust modeling has on the real-world success rates of the motions.
Further results show improved convergence speed when exact Hessians are used over commonly used Hessian approximations such as Gauss-Newton or L-BFGS.

Overall, we obtain a method that enables the computation of robust assembly motions that can directly be executed on real robots without necessity for further tuning, even in the presence of substantial model-reality mismatches. The approach is applicable to any rigid body assembly problem, where all shapes are represented through unions of convex polytopes. The optimal control formulation utilizes model information efficiently, through a differentiable physics simulation formulation which provides up to second-order derivatives. This enables improved sample-efficiency compared to derivative-free methods. The optimal control formulation also allows one to easily add additional constraints in an intuitive way, e.g., one could add bounds on velocities, accelerations or forces.

\subsection{Contribution}
The main contributions of this paper are as follows:
\begin{itemize}
	\item[$\bullet$] A modified formulation of signed distance functions (SDF) as growth distances that closely approximate the Euclidean SDF and a corresponding gradient approximation that is used as contact normal vector.
		\item[$\bullet$]An efficient implementation for SDF and contact normals and corresponding first- and second-order derivatives using the high-performance interior-point solver \texttt{HPIPM} \cite{hpipm} and the linear algebra library \texttt{BLASFEO} \cite{blasfeo}, and a interface that allows one to use these computations in the optimal control library \texttt{CasADi} \cite{casadi}.
	\item[$\bullet$] A multi-scenario-based robust optimal control formulation for assembly planning that ensures successfully executable motions on real robotic systems.
	\item[$\bullet$] Real-world experiments that investigate how smoothing of the contact simulation and robustness parameters of the optimal control problem affect success rates on the real system.
	\item[$\bullet$] Numerical results that illustrate the impact of different smoothing schemes on the convergence behavior of the algorithm and the benefit of using exact Hessians over commonly used approximations such as Gauss-Newton or L-BFGS.
\end{itemize}

\subsection{Notation}
To stack vectors, we compactly write $(x_{1},\dots,x_{m}) = [x_{1}^{\top},\dots,x_{m}^{\top}]^{\top}$ for $x_{1} \in \mathbb{R}^{n_{1}}, \dots, x_{m} \in \mathbb{R}^{n_{m}}$. Furthermore, $\bm{0} = (0,\dots,0)$ and $\bm{1} = (1,\dots,1)$ are the vectors with all zeros or all ones, respectively, where the corresponding dimension becomes clear from the context. 

For a unit quaternion $\xi \in \Rfour$, $R(\xi)$ is the associated $3 \times 3$ rotation matrix. Given another unit quaternion $\zeta \in \Rfour$, $\xi \otimes\zeta$ denotes quaternion multiplication. We further define the identity position $\bar{\rho}_{\mathrm{id}} = (0,0,0)$, the identity quaternion $\bar{\xi}_{\mathrm{id}} = (1,0,0,0)$ and the analogous identity pose $\bar{q}_{\mathrm{id}} = (\bar{\rho}_{\mathrm{id}},\bar{\xi}_{\mathrm{id}})$.

For a multivariate function $f: \mathbb{R}^{n} \times  \mathbb{R}^{m} \rightarrow \mathbb{R}^{l}$, total derivatives are compactly denoted by $\mathrm{D}_{x}f(x,y) = \frac{\mathrm{d}f(x,y)}{\mathrm{d} x} \in \mathbb{R}^{l\times n}$ and partial derivatives are compactly denoted by $\partial_{x}f(x,y) = \frac{\partial f(x,y)}{\partial x} \in \mathbb{R}^{l\times n}$. For a seed $s \in \mathbb{R}^{l}$ second-order total directional derivatives are written as
\begin{equation*}
	\langle s, \, \mathrm{D}^2_{x,y} f(x,y) \rangle = \sum_{i = 1}^{l} s_i \frac{\mathrm{d}^2 f_i(x,y)}{\mathrm{d} x \, \mathrm{d} y} \in \mathbb{R}^{n\times m}.
\end{equation*}
If it is clear from the context, we use the shorthand notation $\langle s, \, \mathrm{D}^2_{x,y} f \rangle$. This notation is equally adapted for second-order partial directional derivatives, thus, we write
\begin{equation*}
	\langle s, \, \partial^2_{x,y} f(x,y) \rangle = \sum_{i = 1}^{l} s_i \frac{\partial^2 f_i(x,y)}{\partial x \, \partial y} \in \mathbb{R}^{n\times m},
\end{equation*}
and in short $\langle s, \, \partial^2_{x,y} f \rangle$.

Table \ref{table:key_symbols} denotes further key symbols used in this paper. We will write $q \in \mathbb{R}^{\nnq}$, $v \in \mathbb{R}^{\nnv}$ and $x \in \mathbb{R}^{\nnx}$ for improved readability and to emphasize that the proposed method can be easily extended to the case with multiple actuated objects, where in this paper the special case $\nnq = 7$, $\nnv = 6$, $\nnx = 13$ is considered.

\subsection{Outline}
This paper is organized as follows. Section \ref{sec:relatedwork} discusses related work on differentiable collision detection, differentiable physics simulation and corresponding planning and control algorithms. In Section \ref{sec:colldet} the proposed optimization-based collision detection formulation, a corresponding contact normal approximation and derivation of first- and second-order derivatives are introduced. The considered dynamical system for contact simulation and corresponding relaxation approaches are discussed in Section \ref{sec:diffphys}. Section \ref{sec:ocp} introduces the optimal control problem formulation for robust assembly planning. Finally, experiments on real hardware and in simulation are conducted in Section \ref{sec:results}, highlighting the properties of the considered methodology.

\newcolumntype{Y}{>{\raggedleft\arraybackslash}X}
\begin{table}[t]
	\centering
	\caption{Overview of key symbols.}
	\label{table:key_symbols}
	\vspace{-3.8pt}
	\begin{tabularx}{\columnwidth}{@{}l@{\hspace{6pt}}l@{\hspace{-12pt}}r@{}}
		\hline \hline
		\textbf{Symbol} & \textbf{Description} & \textbf{First introduced}\\
		\hline \hline
		$\mathbb{R}_{\geq0}$ & Set of nonnegative real numbers& Section \ref{sec:colldet} \\
		$\rho \in \mathbb{R}^{3}$ & Translational position & Section \ref{sec:colldet} \\
		$\xi \in \mathbb{R}^{4}$ & Unit quaternion orientation & Section \ref{sec:colldet} \\
		$q = (\rho,\xi) \in \mathbb{R}^{7}$ & Rigid body pose & Section \ref{sec:colldet} \\
		$\vert \Pi \vert \in \mathbb{N}$ & Number of considered collision pairs & Section \ref{sec:colldet} \\
		$\tau \in \mathbb{R}_{\geq0}$ & Smoothing parameter of collision detection & Section \ref{sec:colldet} \\
		$\Phi_\tau \in \mathbb{R}$ & Signed distance function (SDF) & Section \ref{sec:colldet} \\
		$n_\tau \in \mathbb{R}^{7}$ & Contact normal vector & Section \ref{sec:colldet} \\
		$w_\tau \in \mathbb{R}^{8}$ & Contact information vector & Section \ref{sec:colldet} \\
		\hline
		$\nu \in \mathbb{R}^{3}$ & Translational velocity & Section \ref{sec:diffphys} \\
		$\omega \in \mathbb{R}^{3}$ & Rotational velocity & Section \ref{sec:diffphys} \\
		$v = (\nu,\omega) \in \mathbb{R}^{6}$ & Rigid body velocity & Section \ref{sec:diffphys} \\
		$x = (q,v) \in \mathbb{R}^{13}$ & Rigid body state & Section \ref{sec:diffphys} \\
		$\sigma \in \mathbb{R}_{\geq0}$ & Smoothing parameter of contact resolution & Section \ref{sec:diffphys} \\
		$M \in \mathbb{R}^{6\times 6} $ & Mass matrix & Section \ref{sec:diffphys} \\
		$Q(q) \in \mathbb{R}^{7\times 6} $ & Kinematic map & Section \ref{sec:diffphys} \\
		$\lambda_{\mathrm{n}} \in \mathbb{R}_{\geq0}$ & Contact force magnitude & Section \ref{sec:diffphys} \\
		$\Delta t \in \mathbb{R}_{\geq0}$ &Time step & Section \ref{sec:diffphys} \\
		$H_{\sigma,\tau}(\dots)$ &Contact dynamics equality conditions & Section \ref{sec:diffphys} \\
		$G_{\sigma,\tau}(\dots)$ &Contact dynamics inequality conditions & Section \ref{sec:diffphys} \\
		\hline
		$x_{\mathrm{r}}$ &Reference state & Section \ref{sec:ocp} \\
		$x_{\mathrm{c}}$ &Compliant state & Section \ref{sec:ocp} \\
		$N \in \mathbb{N}$ &Optimization horizon & Section \ref{sec:ocp} \\
		$\nns\in \mathbb{N}$ &Number of scenarios & Section \ref{sec:ocp} \\
		$\hat{q}^{(l)}$ &Offset pose for tracking & Section \ref{sec:ocp} \\
		$\hat{\delta}\in \mathbb{R}_{\geq0}$ & Norm of positional offset & Section \ref{sec:ocp} \\
		$n_{\mathrm{hom}}\in \mathbb{N}$ & Total number of homotopy iterations & Section \ref{sec:ocp} \\
		\hline \hline
	\end{tabularx}
\end{table}

\section{Related Work} \label{sec:relatedwork}
We split the discussion of related work into three parts, the first one on differentiable collision detection, the second one on differentiable physics simulation and the last one on planning and control methods that utilize these simulations to solve contact-rich tasks.

\subsection{Differentiable Collision Detection}
Collision detection in robotics is classically addressed by the Gilbert-Johnson-Keerthi algorithm (GJK) \cite{Gilbert1988}, which determines a separation distance between two convex meshes, and the expanding polytope algorithm (EPA), which is an adapted GJK-variant for determining the penetration distance in case of collision. GJK originally solves the collision detection problem using a simplex method. Though this method exists since thirty years it is still frequently used in many physic engines \cite{Todorov2012Mujoco, SimpleSimulator}. Recent work has shown that the original GJK algorithm is a special case of the Frank-Wolfe algorithm, a first-order optimization method. By combining the Frank-Wolfe interpretation with acceleration methods for gradient descent optimization, a modern GJK implementation with improved efficiency has been developed in \cite{Montaut2024}.

Given a fast method for collision detection, derivatives can be approximately constructed by finite-differences or randomized smoothing \cite{Montaut2023}. This requires many evaluations of the nominal function and unavoidably results in noise on the derivatives which can only be decreased by using more evaluations which becomes computationally expensive. Additionally, the convergence of optimization procedures can be affected by derivative information that does not match the nominal function evaluation, which is the case since the nominal GJK solution is nonsmooth while the derivatives correspond to an implicitly constructed smoothed variant \cite{Pang2023}.

A second approach that closely resembles the method used in this paper, is interpreting collision detection as a convex quadratic or linear program. In this case, interior-point methods can be used to solve for signed distances. Interior-point methods solve a sequence of smoothed problems, where the solution to each smoothed problem becomes differentiable under mild assumptions \cite{Tracy2023, Tracy2024}. For collision detection, this observation was used in conjunction with so-called growth distances \cite{Ong1996}. The great benefit of this method is that it provides a smoothed nominal evaluation of the SDF and by application of the implicit function theorem to its optimality conditions also the corresponding exact first- and second-order derivatives.

Alternatively, one can also use smooth approximations to the maximum operator such as the \texttt{LogSumExp} or \texttt{softmax} operator to obtain smooth analytic point-polytope distances \cite{Yang2025contactsdf}. In \cite{Montaut2023} the \texttt{softmax} approach is extended to polytope-polytope distances by combining it with the GJK algorithm. 

\subsection{Differentiable Physics Simulation}
To solve the contact NCP, various approximation methods and accompanying solver choices have been introduced, each with the intention to enable a reliable and efficient solution of simulation steps. A structured overview and discussion is given in \cite{Lelidec2024contactmodels}. We focus in the following on work that aims at providing derivatives to these contact simulation approaches. To this end, we distinguish four approaches.

In the first case, a method for solving physics simulation steps is combined with finite-differencing or randomized smoothing to obtain derivatives for a nondifferentiable nominal evaluation model. As already discussed for the case of collision detection, this can be prohibitive for optimization approaches where precise convergence should be obtained since nominal function evaluations do not match to the determined derivatives. Even if the nominal problem would be smooth, derivatives obtained this way are only noisy approximations of the exact derivatives, which can harm convergence behavior. These derivatives can however be meaningful for reinforcement learning or model predictive control methods where the optimization procedure is not applied until convergence is achieved. The physics engine \texttt{MuJoCo} \cite{Todorov2012Mujoco} provides first-order derivatives by finite differencing. In \cite{Lelidic2024randomizedsmoothing} randomized smoothing is utilized to differentiate through nonsmooth contact dynamics.

Another class of differentiable physics simulation methods solves for a nonsmooth simulation step and then differentiates with the current active set fixed. That is assuming active contact points remain active and contact points remain on current faces of the geometric shapes. In this case one has a locally smooth model for which exact derivatives can be obtained. This is done in \cite{Werling2021Nimble} for an LCP and in \cite{Lelidec2025} for the original NCP formulation. However, these derivatives do not contain information that switching the active set of contact points or active geometric surfaces could help to solve the underlying task, thus preventing the optimization solver from finding a descent direction. To this end, \cite{Werling2021Nimble} proposes a heuristic to obtain what is referred to as complementarity-aware gradients, meaning derivatives that give information about a possible active set change. Apart from simple examples, it is unclear if such a philosophy can be used to solve complex problems.

Modern implementations of differentiable physics engines use the frameworks \texttt{JAX} and \texttt{Warp} to build computational graphs that are auto-differentiable and that can be efficiently evaluated on GPUs. For \texttt{MuJoCo}, accelerated versions using both frameworks are available and under active development. Differentiating the physic simulation steps of these simulators boils down to differentiating every operation of an iterative solver used to solve the collision detection and contact resolution problem. Thus, in this approach the computational cost for derivative evaluation is directly proportional to the computational cost of the simulation step calculation.

The fourth method to obtain a differentiable physics simulation is to determine a smooth approximation to the nominal NCP evaluation and then to determine the corresponding exact derivatives. This approach is proposed in \texttt{Dojo} \cite{Howell2023Dojo}, where a custom interior-point type solver is used to solve the NCP up to a given smoothing parameter. \texttt{Dojo} does however not consider smoothing of the collision detection contribution. This approach is also used in this paper, whereby we put particular focus on obtaining a smoothing collision detection evaluation and to include it in the physics simulation model.

\subsection{Planning and Control for Contact-Rich Manipulation}
Model-based planning and control approaches can mostly be split into the three categories of sampling-based, reinforcement learning and optimal control methods. Where one can additionally differ between gradient-free reinforcement learning where no derivative information of the dynamic model is utilized and gradient-based reinforcement learning where such derivatives are used for policy optimization.

Impressive results have been accomplished for bi- and quadruped locomotion as well as dexterous manipulation tasks such as pushing objects with robotic arms or in-hand reorientation of objects using actuated multi-finger hands. Without the usage of derivative information in sampling-based approaches \cite{Howell2022predictivesampling, AlvarezPadilla2024realtimewholebody} and gradient-free reinforcement learning \cite{Li2025RL, Seo2025FastTD3, KimKaist2025}, as well as with usage of first-order derivatives in model-based reinforcement learning \cite{Xu2022} and model predictive control methods \cite{Tassa2012, Kim2024KaistMPC, Jiang2024, Kurtz2025}. Also combinations of sampling-based, RL and MPC methods are utilized to succeed in solving the mentioned tasks \cite{Levine13,Pang2023,Liu2025Opt2Skill}.

For assembly planning, mainly work which utilizes sampling-based approaches and gradient-free reinforcement learning exists, illustrating the lack of work that utilizes differentiable physics simulations for this problem class. In \cite{Vuong2023, Lee2024} robust peg assembly is achieved via gradient-free reinforcement learning under excessive use of simulation steps. In \cite{Wirnshofer2018} similar problems are solved through random sampling of reference trajectories which are tracked by multiple domain randomized compliant trajectories through a Cartesian impedance law.
To randomize compliant trajectories, we use the domain randomization approach of \cite{Mordatch2015}, which was originally utilized to robustify a controller for biped walking.

\section{Smooth Differentiable Collision Detection} \label{sec:colldet}
Here, we derive a modified variant of an established growth distance \cite{Ong1996} that ensures well-scaled contact-implicit optimal control problems. This growth distance is a signed distance formulation between convex polytopes. In Section \ref{sec:colldet:optbasedcolldec}, we state the nominal formulation and discuss its differentiability. In Section \ref{sec:colldet:gradientapprox}, we propose an approximation to the gradient of the SDF that will be used as contact normal vector for the contact resolution problem. In Section \ref{sec:colldet:derivatives}, we derive first- and second-order derivatives for the SDF and the corresponding contact normal vector and discuss their efficient implementation. In Section \ref{sec:colldet:rescaling}, we propose a heuristic approach to reliably solve the nominal problem by introducing scaled variants with the same solution that are solved if the numerical solver fails on the original problem.
\subsection{Optimization-based Collision Detection} \label{sec:colldet:optbasedcolldec}
\begin{figure}
	\centering
	\includegraphics[width=4.3cm]{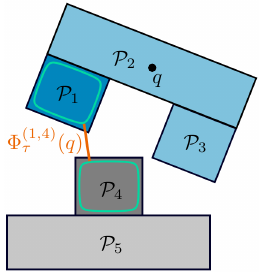}
	\caption{Illustration of the type of assembly problem addressed in this paper. For this specific example, we have $\nnact = 3$ and $\nnenv = 2$.}
	\label{fig:problem_description}
\end{figure}

Our goal is to simulate the contact interactions of one actuated rigid object with another rigid object fixed in the environment. This emulates assembling one object onto another on a fixed position, a common process in automated industrial production lines. To this end, the actuated object consists of $\nnact \in \NN$ polytopes and the environment consists of $\nnenv \in \NN$ polytopes, cf. Figure \ref{fig:problem_description}. The three-dimensional polytopes are represented in a halfspace representation given by
\begin{equation*}
	\mathcal{P}_{i} = \{p \in \mathbb{R}^{3}\ \vert \ G_{i}p \leq h_{i}\}, \quad i=1,\dots, \nnact + \nnenv,
\end{equation*}
where $G_{i} \in \mathbb{R}^{\nhi \times 3}$ and $h_{i} \in \mathbb{R}^{\nhi}$. We make the following assumptions on the polytopes that are without loss of practical applicability.
\begin{assumption} \label{assumption_1}
	Let $\mathcal{P}_{i}$ be bounded convex polytopes that have nonempty interior. Let the matrix $G_{i}$ be constructed such that its rows are vectors with norm unity. Let $h_{i} > \bm{0}$, i.e., the origin is contained in the interior of the polytopes.
\end{assumption}

To simulate the contact interactions between the actuated and environmental polytopes, we consider a set of relevant contact pairs
\begin{equation*}
	\Pi \subset \{ (i,j) \mid i = 1,\dots,\nnact, \ j = \nnact + 1,\dots, \nnact + \nnenv \}.
\end{equation*}
For each polytope, we associate positional offsets, namely $\rhoi\in \Rthree$ for the translational position and the unit quaternion $\xii \in \Rfour$ for the orientation. These offsets determine the position of the polytopes with respect to a parent coordinate frame. In case of the polytopes making up the environment this will be the world frame, in case of the actuated polytopes this is a frame defined by a common degree of freedom (DoF) \mbox{$q = (\rhod, \xid) \in \mathbb{R}^{7} $}, where $\rhod \in \Rthree$ corresponds to the translational position and $\xid \in \Rfour$ corresponds to a unit quaternion orientation of the actuated object in the world frame.
\begin{assumption} \label{assumption_2}
	Let the offsets $\rhoi$ and $\xii$ for $i = 1,\dots,\nnact$ be constructed such that $\rhod$ corresponds to the center of mass (CoM) of the actuated object.
\end{assumption}
The translational and rotational position of the actuated polytopes in the world frame is retained through
\begin{align*}
\bar{\rho}_{a, i}(q) &= \rhod + R(\xid) \rhoi, \\
\bar{\xi}_{a, i}(q) &= \xid \otimes \xii,
\end{align*}
where $R(\cdot)$ is the $3 \times 3$ rotation matrix associated with an unit quaternion and $\otimes$ denotes quaternion multiplication. For the remainder of this section, we focus on a fixed distance pair $(i,j) \in \Pi$. We define the SDF for this pair by
\begin{mini}
	{p,\alpha}{2\alpha \phantom{aaaaaaaaaaaaaaaaaaaaaaaaaaaaaaa}}{\label{sdfopt}}{\Phi_{0}(q) = }
	\addConstraint{G_{i} R(\bar{\xi}_{a, i}(q))^{\top}(p-\bar{\rho}_{a, i}(q))}{\leq h_{i} + \alpha \bm{1},}
	\addConstraint{G_{j} R( \xiej)^{\top}(p-\rhoej)}{\leq h_{j} + \alpha \bm{1},}
\end{mini}
where smoothed variants $\Phi_{\tau}(q)$ for $\tau > 0$ will be introduced in the following. A similar formulation was originally devised in \cite{Ong1996} and termed as growth distance as it determines the smallest scaling factor $\alpha \in \mathbb{R}$ by which two polytopes have to be altered such that there is a common point $p \in \Rthree$ in both scaled shapes. We note a slight but for our purposes crucial modification to previous work: instead of scaling the polytopes in a form-stable manner through multiplying $\alpha$ to the right-hand side of the halfspace representations ($\alpha h_i$ and $\alpha h_j$ as in \cite{Ong1996}), we apply a uniform scaling in all directions through the term $\alpha\bm{1}$. This proved to be useful to retain well-scaled optimization problems when this SDF is used within contact-implicit optimal control. Additionally, a factor of two is introduced into the cost function in \eqref{sdfopt} to obtain a more accurate approximation of the corresponding Euclidean distance. This factor is necessary because the scaling is applied to both polytopes.

We construct a simple point-polytope distance in two dimensions based on the above introduced signed distance function that will be used for visualization purposes.
\begin{example}
Let $\rho^{\mathrm{2d}} \in \mathbb{R}^{2}$, $G^{\mathrm{2d}} \in \mathbb{R}^{n_{\mathrm{h}} \times 2}$, $h^{\mathrm{2d}} \in \mathbb{R}^{n_{\mathrm{h}}}$. By transferring formulation \eqref{sdfopt} to compute the distance of $\rho^{\mathrm{2d}}$ and the fixed polytope $\mathcal{P}^{\mathrm{2d}} = \{p^{\mathrm{2d}} \in \mathbb{R}^{2}\ \vert \ G^{\mathrm{2d}}p^{\mathrm{2d}} \leq h^{\mathrm{2d}}\}$, we obtain the formulation
\begin{mini*}
	{p^{\mathrm{2d}},\alpha}{\alpha \phantom{aaaaaaaaaa}}{}{\Phi^{\mathrm{2d}}_{0}(\rho^{\mathrm{2d}}) = }
	\addConstraint{p^{\mathrm{2d}} = \rho^{\mathrm{2d}},}
	\addConstraint{G^{\mathrm{2d}}p^{\mathrm{2d}}}{\leq h^{\mathrm{2d}} + \alpha \bm{1}.}
\end{mini*}
Here, no factor of two is introduced into the cost function since only one polytope is scaled. All following general derivations for $\Phi_{0}(q)$ may be equivalently done for $\Phi^{\mathrm{2d}}_{0}(\rho^{\mathrm{2d}})$.
\end{example}

\begin{figure}
	\centering
	\includegraphics{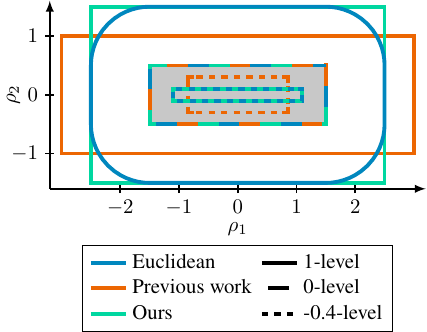}
	\caption{The \{1,0,-0.4\}-level lines for different optimization-based distance formulations for the distance between a point and the gray cuboid. Previous work refers to \cite{Ong1996,Tracy2023}. The 0-level lines coincide for all distance formulations.}
	\label{fig:distancecomp}
\end{figure}

Figure \ref{fig:distancecomp} shows different level lines for point-polytope distances with a two-dimensional cuboid. The Euclidean distance, the previously established growth distance \cite{Ong1996} and our formulation are considered. The 0-level line coincides for all formulations but during the trajectory optimization process values larger and smaller than zero are also crucially relevant, thus, well-scaled distance formulations are desirable. Using Euclidean distances is the gold standard but signed Euclidean distances may be only formulated as nonconvex optimization problems \cite{zhang2021}, making it challenging to efficiently solve these distance formulations. In contrast to previous work, the here proposed growth distance exactly matches the Euclidean distance for negative values and closely approximates it for positive values.

Now, to calculate the SDF $\Phi_{0}(q)$, we consider the Karush-Kuhn-Tucker (KKT) system of the corresponding LP \eqref{sdfopt}. To this end, denote the number of inequality conditions by $n_{\mathrm{g}} = \nhi + \nhj$. We then represent the inequality constraints in compact notation and define
\begin{equation*}
A(q) = \begin{pmatrix}
	G_{i}  R(\bar{\xi}_{a, i}(q))^{\top} & -\bm{1} \\
	G_{j} R( \xiej)^{\top} & -\bm{1}
\end{pmatrix} \in \mathbb{R}^{n_{\mathrm{g}} \times 3}
\end{equation*}
and
\begin{equation*}
	b(q) = \begin{pmatrix}
		h_{i} + G_{i}R(\bar{\xi}_{a, i}(q))^{\top}\bar{\rho}_{a, i}(q)\\
			h_{j} + G_{j} R( \xiej)^{\top}\rhoej
	\end{pmatrix} \in \mathbb{R}^{n_{\mathrm{g}}}.
\end{equation*}
Denoting the primal variables by $z = (p,\alpha) \in \mathbb{R}^{4}$, the dual variables by $\lambda \in \mathbb{R}^{ n_{\mathrm{g}}}$ and the combined primal-dual variables as $\gamma = (z,\lambda) \in \mathbb{R}^{4 + n_{\mathrm{g}}}$, solving the problem corresponds to solving the KKT system for $\tau = 0$ which is given by
\begin{equation} \label{KKTsystem}
F_{\tau}(\gamma,q) = \begin{pmatrix}
c + A(q)^{\top} \lambda \\
\Lambda ( b(q) - A(q)z) - \bm{1}\tau
\end{pmatrix} = 0,
\end{equation}
under the additional constraints
\begin{equation}  \label{KKTconstraints}
\lambda \geq 0, \ b(q) - A(q)z \geq 0.
\end{equation}
Here $\Lambda = \diag(\lambda)$ is the diagonal matrix constructed from $\lambda$ and $c = (\bm{0},2)$ is the cost vector for \eqref{sdfopt}. Interior-point methods solve \eqref{KKTsystem} by applying Newton steps to the system of equations starting with a large $\tau\gg0$ and then consecutively decreasing this value until a solution for the exact system with $\tau = 0$ is found up to a user-defined tolerance. However, solving the problem exactly makes the distance definition nondifferentiable. In \cite{Dietz2025} it was shown that under \mbox{Assumption \ref{assumption_1}} solving the KKT system up to any $\tau > 0$ ensures existence of a unique primal-dual solution $\gamma$. This allows one to apply the implicit function theorem to \eqref{KKTsystem} to obtain corresponding derivatives of arbitrary order.

We therefore introduce the notation $z_{\tau}(q) = (p_{\tau}(q),\alpha_{\tau}(q))$ and further $\gamma_{\tau}(q) = (z_{\tau}(q),\lambda_{\tau}(q))$ for the unique primal-dual solution of \eqref{KKTsystem} and \eqref{KKTconstraints} for given $\tau > 0$ and $q$. This enables the following definition of a smooth SDF approximation given by
\begin{equation} \label{smoothdistfunc}
	\Phi_{\tau}(q) = \alpha_{\tau}(q).
\end{equation}
This SDF approximation has the following properties.
\begin{proposition} \label{prop1} Let Assumption 1 hold. Then the SDF approximation $\Phi_{\tau}(q)$ for $\tau > 0$ is a well-defined infinitely differentiable function. For $\tau \rightarrow 0$, the smooth SDF $\Phi_{\tau}(q)$ converges to the nonsmooth SDF $\Phi_{0}(q)$. It holds $\Phi_{\tau}(q) \geq \Phi_{0}(q)$.
\end{proposition}
To show this, one can adapt the proof of \cite{Dietz2025} to the three-dimensional case and the proposed SDF modification. The key requirement is to ensure that $A(q)$ has rank three which is the case for polytopes with nonempty interior. 

The characteristic that the considered SDF approximation is an upper bound for the exact signed distance is crucial, as this ensures that the feasible space of configurations is not decreased. Thus, we have
\begin{equation*}
	\{ q \in \mathbb{R}^{\nnq} \mid  \Phi_{0}(q) \geq 0 \} \subset \{ q \in \mathbb{R}^{\nnq} \mid  \Phi_{\tau}(q) \geq 0 \}.
\end{equation*}
This is of particular importance when considering assembly problems involving tight clearances, since for such problems the feasible configuration space is highly confined. Using a SDF approximation that underestimates the exact distance would then quickly lead to infeasible problem formulations, for which no valid solution trajectory exists.

\subsection{SDF Gradient Approximation} \label{sec:colldet:gradientapprox}
For physics simulation, one requires a contact normal vector for each touching contact pair that defines the direction in which contact forces act. Classically, one uses the gradient of the SDF \cite{Brogliato2016}. However, since we aim to use second-order optimization methods together with the considered contact simulation formulation, defining the contact normal vector as the SDF gradient results in third-order derivatives of the SDF that have to be passed to the numerical solver in contact-implicit optimal control. Therefore, we propose an approximation to the gradient of the smooth SDF $\Phi_{\tau}(q)$ that is a function of the primal-dual variables of the distance problem. Differentiating the dynamical system twice will then only result in the requirement of differentiating the primal-dual variables of the distance problem twice with respect to the object poses $q$. The contact normal approximation is motivated by the following observation as discussed in \cite{Dietz2025}. Since the SDF \eqref{sdfopt} is a parametric LP in $q$, we denote by $\bar{Z}_{0}(q)$ the set of all primal optimal solutions and by $\bar{\Lambda}_{0}(q)$ the set of all corresponding dual optimal solutions. Whenever one of these sets is not a singleton set for a given $q$, the SDF $\Phi_{0}(q)$ is nondifferentiable. Further, let $g(z,q) = A(q)z-b(q)$ compactly capture the inequality constraints of the distance problem \eqref{sdfopt}. Now directional derivatives for the nonsmooth SDF $\Phi_{0}(q)$ are given by
\begin{equation} \label{dirderv}
		\lim_{h\rightarrow 0}\frac{\Phi_{0}(q+hd) - \Phi_{0}(q)}{h\Vert d \Vert} 
		 = \min_{z \in \bar{Z}_{0}(q)}\max_{\lambda \in \bar{\Lambda}_{0}(q)} d^{\top} \delq g(z,q)^{\top} \lambda ,
\end{equation}
for direction $d\in \mathbb{R}^{n_{q}}$, see \cite{Hogan1973}. This motivates us to define the gradient approximation for the smooth SDF as
\begin{equation} \label{contactvectordef}
	n_{\tau}(\gamma_{\tau}(q),q) = \delq g(z_{\tau}(q),q)^{\top} \lambda_{\tau}(q).
\end{equation}
In compact form we also write $n_{\tau}(q) \coloneqq n_{\tau}(\gamma_{\tau}(q),q)$. 

\begin{figure}
	\centering
	\includegraphics[width=1\linewidth]{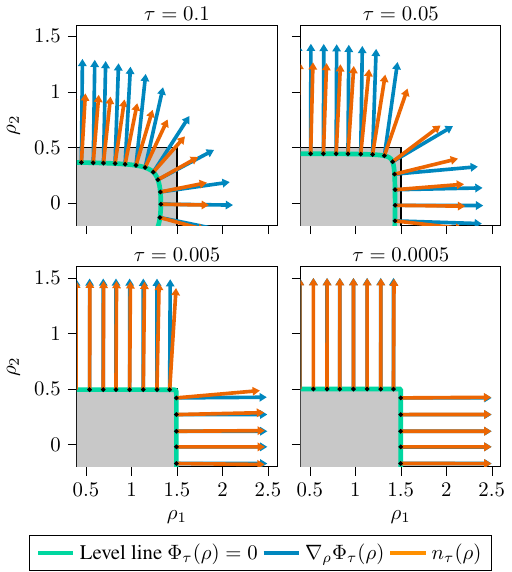}
	\caption{Exact gradients $\nabla_{\rho} \Phi_\tau(\rho)$ and the proposed gradient approximation $n_\tau(\rho)$ for a point-polytope distance in two dimensions. Four different values of $\tau$ are considered, the behavior that for $\tau \rightarrow 0$ the approximation converges to the exact gradient can be observed.}
	\label{fig:normalvis}
\end{figure}

This gradient approximation naturally has the property that whenever $\nabla_{q} \Phi_{0}(q)$ classically exists, i.e., $\bar{Z}_{0}(q)$ and $\bar{\Lambda}_{0}(q)$ are singletons, then $n_{\tau}(q)$ converges to $\nabla_{q} \Phi_{0}(q)$ for $\tau \rightarrow 0$. This is illustrated in Figure \ref{fig:normalvis} for the two-dimensional point-polytope distance that was already considered in Figure \ref{fig:distancecomp}. One can observe that $n_{\tau}(q)$ is generally a close approximation of $\nabla_{q} \Phi_{0}(q)$ that can be made arbitrarily accurate for $\tau \rightarrow 0$. Furthermore, the exact and approximate gradient attain norm unity for $\tau \rightarrow 0$, which ensures the well-scaledness of the optimization problems in which the gradient approximation will be utilized, as already mentioned in Section \ref{sec:colldet:optbasedcolldec}. If one would use the growth distance formulation introduced in previous work \cite{Ong1996,Tracy2023} (the orange level lines in Figure \ref{fig:distancecomp}), one would obtain exact and approximate gradients with varying norms depending on the shape of the considered polytopes.

If the primal or dual optimal set is not a singleton, an interior-point method converges to the primal-dual solution corresponding to the analytic center of these sets \cite{Bertsekas2016}. Consequentially, $n_{\tau}(q)$ then converges to the subgradient for this particular primal-dual solution.

\begin{remark}
The considered SDF gradient approximation can also be interpreted as the exact gradient for the objective value of the associated barrier problem to the distance LP \eqref{sdfopt}. I.e., consider the barrier function
\begin{equation*}
B_{\tau}(z,q) = c\tsp z - \tau \sum_{i= 1}^{\nng} \log(-g_i(z,q)),
\end{equation*}
and define the alternative smooth SDF by
\begin{align}
	\tilde{\Phi}_{\tau}(q) &= \min_{z} B_{\tau}(z,q) \label{barrierprob} \\
	&= B_{\tau}(z_\tau(q),q), \notag
\end{align}
where the second equality follows from the definitions made in Section \ref{sec:colldet:optbasedcolldec} since \eqref{KKTsystem} and \eqref{KKTconstraints} denote the optimality conditions of the barrier problem. One can check that now it holds
\begin{align*}
	\nabla_q \tilde{\Phi}_{\tau}(q) = n_\tau(q),
\end{align*}
since we have by the chain rule
\begin{align*}
	\dq \tilde{\Phi}_{\tau}(q) &= \delq B_{\tau}(z_\tau(q), q) +  \partial_{z} B_{\tau}(z_\tau(q), q) \dq z_\tau(q) \\
	&= \delq B_{\tau}(z_\tau(q), q),
\end{align*}
and the second equality follows since due to optimality of $z_\tau(q)$ for \eqref{barrierprob} it holds $\partial_{z} B_{\tau}(z_\tau(q), q) = 0$. By evaluating the partial derivative of the barrier function one then obtains
\begin{align*}
\delq B_{\tau}(z_\tau(q), q) &= \lambda_{\tau}(q) \delq g(z_{\tau}(q),q) \\
&= n_\tau(q)\tsp.
\end{align*}
\end{remark}

\subsection{First- and Second-Order Derivatives of Collision Detection} \label{sec:colldet:derivatives}
In this section first- and second-order derivatives for the SDF $\Phi_{\tau}(q)$ and the corresponding contact normal vector $n_{\tau}(q)$ are derived. We analytically compute these derivatives to guarantee a high-performance implementation that utilizes problem structures efficiently. To this end, we define the contact information function
\begin{equation} \label{contactinformation}
w_{\tau}(\gamma_{\tau}(q),q) = \begin{pmatrix}
	\alpha_{\tau}(q) \\
	n_{\tau}(\gamma_{\tau}(q),q)
\end{pmatrix} \in \mathbb{R}^{\nnq + 1}.
\end{equation}
This function contains all the contact information that will be used in the contact-rich dynamical system. In the following we drop the arguments and subscripts for the functions $w_{\tau}(\gamma_{\tau}(q),q)$ and $F_{\tau}(\gamma, q)$ for improved readability. We need to provide the Jacobian $\dq w$ and the directional Hessian $\langle  \sw \, , \,  \dqtwo w \rangle$, for a seed $\sw \in \mathbb{R}^{\nnq +1}$, which are used by Newton-based numerical optimization solvers.

For the Jacobian of \eqref{contactinformation} we have by the chain rule
\begin{equation} \label{firstderw}
\dq w = \delgamma w \, \dq \gamma + \delq w.
\end{equation}
The Jacobian of the primal-dual variables $\dq \gamma$ is obtained from the implicit function theorem (IFT) by solving the linear system
\begin{equation} \label{firstderF}
\delgamma F	 \, \dq \gamma = -\delq F,
\end{equation}
where the invertibility of $\delgamma F$ is directly ensured by the uniqueness of $\gamma_\tau(q)$, i.e., a rank deficient $\delgamma F$ would imply a nonunqiue primal-dual solution.

To obtain directional second-derivatives of \eqref{contactinformation}, we need to differentiate $\sw \tsp w$ twice w.r.t. $q$. To this end, we first obtain the Jacobian of $\sw\tsp w$ by multiplying \eqref{firstderF} with the seed $\sw$ and obtain
\begin{equation*}
	\sw\tsp \dq w = \sw\tsp \delgamma w \, \dq \gamma +\sw\tsp  \delq w.
\end{equation*}
Differentiating a second time with respect to $q$ using the chain rule results in
\begin{equation} \label{secondderw}
	\begin{split}
	\langle  \sw \, , \,  \dqtwo w \rangle &= \langle  \sw \, , \, \delqtwo w \rangle \\
	&+ \langle  \sw \, , \,  \delqgamma w \rangle \dq \gamma \\
	&+ \dq \gamma \tsp \langle  \sw \, , \,  \delgammaq w \rangle \\
	&+ \dq \gamma \tsp \langle  \sw \, , \,  \delgammatwo w \rangle \dq \gamma \\
	&+ \langle  s_\gamma \, , \, \dqtwo\gamma \rangle,
\end{split}
\end{equation}
where we obtained a new seed $\sgamma \in \mathbb{R}^{\nngamma}$ as
\begin{equation*}
\sgamma = \delgamma w \tsp \sw.
\end{equation*}
It remains to determine the directional Hessian of the primal-dual variables $\langle  s_\gamma \, , \, \dqtwo\gamma \rangle$. To this end, we interpret \eqref{firstderF} as a parametric linear system and apply the IFT one more time. A detailed derivation is given in Appendix \ref{appendix:A}. One then first solves the adjoint equation
\begin{equation} \label{secondderadjeq}
\delgamma F \tsp	 \, r_\gamma = -s_\gamma,
\end{equation}
to obtain $r_\gamma \in \mathbb{R}^{\nngamma}$. Then we have
\begin{equation} \label{secondderfF}
	\begin{split}
	\langle  s_\gamma \, , \, \dqtwo\gamma \rangle &= \langle  r_\gamma \, , \, \delqtwo F \rangle \\
	&+ \langle  r_\gamma \, , \,  \delqgamma F \rangle \dq \gamma \\
	&+ \dq \gamma \tsp \langle  r_\gamma \, , \,  \delgammaq F \rangle \\
	&+ \dq \gamma \tsp \langle  r_\gamma \, , \,  \delgammatwo F \rangle \dq \gamma.
	\end{split}
\end{equation}

To obtain a highly efficient implementation, many quantities evaluated during construction of first-order derivatives in \eqref{firstderw} and \eqref{firstderF} can be reused for the construction of second-order derivatives in \eqref{secondderw}-\eqref{secondderfF}. Notably, \eqref{firstderF} is solved through the construction of a LU decomposition and subsequent backsolving where the LU decomposition is then reused for \eqref{secondderadjeq}. \\
Furthermore, we notice the following relation between the contact normal vector and the KKT system. For the selection vector $s_{\mathrm{s}} = (\bm{0},\bm{1})$, we have
\begin{equation*}
	n_{\tau}(\gamma_{\tau}(q), q) =  - \delq F_{\tau}(\gamma_{\tau}(q), q) \tsp s_{\mathrm{s}}.
\end{equation*}
Thus, when dividing contact information evaluation in three phases: nominal, first derivative and second derivative evaluation, some derivatives of $F$ with respect to $q$ naturally have already been evaluated in the preceding phase.

\subsection{Reliable solutions for the distance problems} \label{sec:colldet:rescaling}
To obtain solutions to the distance problem \eqref{sdfopt}, we utilize the high-performance interior-point solver \texttt{HPIPM} \cite{hpipm}, which supports the option to solve the problem only up to a fixed barrier parameter $\tau > 0$. However, in rare cases the solver may fail to converge to a solution of \eqref{sdfopt}, as it may run into a primal-dual pair where the KKT matrix becomes ill-conditioned. To ensure that correct contact information can always be passed to the OCP solver, we construct scaled variants of \eqref{sdfopt} from which the solution of the original problem can be retained. The process is as follows.

We determine uniformly drawn random vectors $\zeta_{l} \sim \mathcal{U}[\kappa_{\mathrm{min}},\kappa_{\mathrm{max}}]^{n_{\mathrm{g}}}$ for $l = 1,\dots,n_{\mathrm{tries}}$ with $\kappa_{\mathrm{min}},\kappa_{\mathrm{max}}>0$ and $n_{\mathrm{tries}} \in \mathbb{N}$, where we, e.g., use $\kappa_{\mathrm{min}} =1, \kappa_{\mathrm{max}} = 10$ and $n_{\mathrm{tries}}=20$. After a failed solve attempt of \eqref{sdfopt}, we switch instead to the problem
\begin{mini}
	{z}{c\tsp z}{\label{sdfopt:scaled}}{}
	\addConstraint{\widetilde{A}(q)z \leq \widetilde{b}(q),}
\end{mini}
where
\begin{equation*}
\widetilde{A}(q) = \diag(\zeta_{l})A(q), \quad \widetilde{b}(q) = \diag(\zeta_{l})b(q).
\end{equation*}
This definition implies that for $\zeta_{l} = \bm{1}$, the original problem \eqref{sdfopt} is retained. Solving \eqref{sdfopt:scaled} results in various evolutions of iterates produced by the interior-point solver and, thus, enables converge to a solution if another evolution of iterates ended up in an ill-conditioned KKT matrix.

A primal-dual solution $\widetilde{\gamma} = (\widetilde{z}, \widetilde{\lambda})$ of \eqref{sdfopt:scaled} up to the barrier parameter $\tau > 0$ then satisfies the perturbed KKT conditions analogous to \eqref{KKTsystem} and \eqref{KKTconstraints} given by
\begin{equation} \label{KKTsystem:scaled}
	\begin{split}
	\begin{pmatrix}
			c + \widetilde{A}(q)^{\top} \widetilde{\lambda} \\
			\Lambda ( \widetilde{b}(q) - \widetilde{A}(q)\widetilde{z}) - \bm{1}\tau
		\end{pmatrix} &= 0, \\
		\widetilde{\lambda} \geq 0, \ \widetilde{b}(q) - \widetilde{A}(q)\widetilde{z} &\geq 0.
		\end{split}
\end{equation}
One can then deduce, that if $\widetilde{z}$ and $\widetilde{\lambda}$ satisfy \eqref{KKTsystem:scaled}, then $z = \widetilde{z}$ and $\lambda = \diag(\zeta_{l}^{-1})\widetilde{\lambda}$ satisfy \eqref{KKTsystem} and \eqref{KKTconstraints}, where $\zeta_{l}^{-1} = (\zeta_{l,1}^{-1},\dots,\zeta_{l,n_{\lambda}}^{-1})$. In our case, we have to re-solve less than $1\%$ of the considered distance problems and always obtain a solution after a few attempts on scaled problems.

\section{Differentiable Multi-Contact Physics}\label{sec:diffphys}
We formulate frictionless contact dynamics between rigid bodies with Cartesian coordinates. In Section \ref{sec:diffphys:cnttime} continuous-time dynamics are stated and in Section \ref{sec:diffphys:dsctime} the corresponding discrete-time dynamics are derived by utilizing a common time-stepping scheme.

\subsection{Continuous-Time Contact Dynamics} \label{sec:diffphys:cnttime}
 Continuous-time dynamics of frictionless contact-rich systems combine Newton-Euler equations with complementarity conditions for the determination of contact forces. Additionally, in continuous time, multi impact laws have to be considered that describe how system velocities jump if objects initially make contact. Choosing such a multi impact law is a modeling decision \cite{Brogliato2016}. A dynamical system describing the motion of a single rigid body interacting with a static environment is given by
\begin{subequations} \label{cls}
	\begin{align}
		&\dot{q} = Q(q)v, \label{cls:a}\\
		&M \dot{v} = U + \sum_{\dind \in \Pi} Q(q)\tsp n_{\tau}^{\dind}(q)\lambda_{\mathrm{n}}^{\dind}, \label{cls:b}\\
		&0 \leq  \Phi_{\tau}^{\dind}(q)  \ \perp \ \lambda_{\mathrm{n}}^{\dind} \geq 0, \quad  \dind \in \Pi, \label{cls:c} \\
		&\text{Multi impact law.} \label{cls:f}
	\end{align}
\end{subequations}
The system velocity $v \in \mathbb{R}^{\nnv}$ is integrated into pose space through the kinematic map $Q(q) \in \mathbb{R}^{\nnq\times \nnv}$. It is required due to the presence of quaternion dynamics as quaternions are represented by four-dimensional vectors while the corresponding rotational velocity is given through a three-dimensional vector. The kinematic map is also used to project the contact normal vector to the velocity space. For its definition see, \mbox{e.g., \cite{Gros2015}}. Furthermore, $M \in \mathbb{R}^{\nnv \times \nnv}$ is the positive definite inertia matrix and \mbox{$U \in \mathbb{R}^{\nnv}$} denotes externally applied forces. The system state is denoted compactly by $x = (q,v)$. The contact force magnitudes are denoted $\lambda_{\mathrm{n}}^{(i,j)}\in \mathbb{R}_{\geq0}$.

\subsection{Discrete-Time Contact Dynamics} \label{sec:diffphys:dsctime}
Discrete-time dynamics are derived by applying a time-stepping scheme to \eqref{cls}. For compact notation we define the contact normal mapped to velocity space according to \eqref{cls:b} as 
\begin{equation*}
	\tilde{n}_{\tau}^{\dind}(q) = Q(q)\tsp n_{\tau}^{\dind}(q) \in \mathbb{R}^{\nnv}.
\end{equation*}
 We consider the common choice of an semi-implicit Euler time-stepping method. That is using an implicit Euler step on \eqref{cls:a}, an explicit one on \eqref{cls:b} and an implicit evaluation of the algebraic equation \eqref{cls:c}, whereby the SDF is linearized around the previous state. This scheme was originally introduced by Moreau \cite{Moreau} and corresponds to the optimality conditions of a convex quadratic program. The resulting discrete-time system for the time step $\Delta t> 0$ is given by
\begin{subequations} \label{ts}
	\begin{align}
		&q_{k+1}= q_{k} + \Delta t Q(q_{k+1})v_{k+1}, \label{ts1}\\
		&v_{k+1}= v_{k} + \Delta tM^{-1}\Big(U_{k}+ \sum_{(i,j) \in \Pi} \tilde{n}_{\tau}^{\dind}(\tilde{q}_{k})\lambda_{\mathrm{n},k+1}^{(i,j)} \Big), 
		\\
		\begin{split}
		&0 \leq \Phi_{\tau}^{(i,j)}(\tilde{q}_{k}) + \Delta t \tilde{n}_{\tau}^{\dind}(\tilde{q}_{k})\tsp v_{k+1} \, \\
		&\hspace{3.41cm} \perp \lambda_{\mathrm{n},k+1}^{(i,j)} \geq 0, \quad \forall (i,j) \in \Pi.
	\end{split} \label{ts3}
	\end{align}
\end{subequations} 
We introduce a slight modification of the pose given by $q_k = (\rho_k,\xi_k)$ as we instead use $\tilde{q}_k$ in \eqref{ts} when evaluating the SDF and the contact normal. The modified pose is given by
\begin{equation*}
\tilde{q}_k = (\rho_k,\frac{\xi_k}{\Vert \xi_k \Vert_2}).
\end{equation*}
On the one hand, due to the integration of quaternions in Euler space \eqref{ts1} the quaternions will not maintain norm unity. This error can however be controlled by choosing a small enough $\Delta t$ and was not a main concern during our experiments. More importantly, since we will solve the system \eqref{ts} within a trajectory optimization framework, the optimizer may use pose values in intermediate iterations that are not close to fulfilling \eqref{ts} and we want to avoid evaluating the SDF implementation for quaternion values that are far from norm unity.

Due to the complementarity conditions \eqref{ts3}, the solution of the time-stepping scheme can be nondifferentiable with respect to the previous position and velocity as well as the control input. In this paper, we take the approach of directly transcribing the system \eqref{ts} into the optimal control problem. Thereby, the OCP solver will simultaneously solve for the optimal control input and the system dynamics. However, we still rely on relaxing the complementarity conditions to ensure a differentiable system. To this end, we utilize Scholtes' smoothing or relaxation \cite{Scholtes}, which is a well-established method for complementarity condition relaxation and a choice particularly suitable for use in optimization applications \cite{Nurkanovic2024c}.

By compactly denoting the complementarity conditions \eqref{ts3} by
\begin{equation*}
	0 \leq a_l \, \perp \, b_l \geq 0, \quad l = 1,\dots, \vert \Pi \vert,
\end{equation*}
Scholtes' smoothing is given through the conditions
\begin{equation*}
	0 \leq a_l, \ 0 \leq b_l, \ a_l b_l =\sigma, \quad l = 1,\dots, \vert \Pi \vert,
\end{equation*}
for some $\sigma > 0$ and Scholtes' relaxation is given through
\begin{equation*}
	0 \leq a_l, \ 0 \leq b_l, \ a_l b_l \leq\sigma, \quad l = 1,\dots, \vert \Pi \vert.
\end{equation*}
Since the system \eqref{ts} corresponds to the optimality conditions of a convex quadratic problem, utilizing smoothing implied by interior-point methods is equivalent to applying Scholtes' smoothing to the LCP formulation. Scholtes' relaxation on the other hand, relaxes each complementarity condition individually, and, therefore, enables the OCP solver to use these additional degrees of freedom in the physics simulation to minimize the OCP's cost function. Using Scholtes' relaxation mirrors the idea of Todorov \cite{TodorovOptico} to obtain contact-rich motions from trajectory optimization by first allowing contact forces at a distance that help to achieve the task goal and later aiming to optimize for control inputs such that the system attains positions where these contact forces become actually physical. Since system \eqref{ts} in conjunction with Scholtes' smoothing/relaxation consists for any $\tau > 0$ and $\sigma> 0$ of smooth functions, we can write \eqref{ts} compactly as smooth equality and inequality conditions in the form
\begin{align*}
&H_{\sigma,\tau}(x_{k},x_{k+1},U_{k},\lambda_{\mathrm{n},k}) = \bm{0}, \\
&G_{\sigma,\tau}(x_{k},x_{k+1},U_{k},\lambda_{\mathrm{n},k}) \leq \bm{0}.
\end{align*}
First- and second-order derivatives for this system, apart from SDF and contact normal evaluations, will be derived by automatic differentiation using \texttt{CasADi}.

Overall, we derived a completely smooth physics simulation system, where we smoothed both the collision detection LCP and the contact resolution LCP through interior-point smoothing, which can optionally be replaced for the contact resolution LCP by Scholtes' relaxation method. This smooth approximation of the original nonsmooth physics simulation can be made arbitrarily accurate by letting $\tau \rightarrow 0$ and $\sigma \rightarrow 0$.

\section{Robust Contact-Implicit Optimal Control} \label{sec:ocp}
In this section, we formulate an open-loop optimal control problem that generates assembly motions which can be reliably executed on real robotic systems. To this end, we utilize a multi-scenario-based formulation that optimizes for a robust reference trajectory. The reference trajectory is optimized such that successful task completion is achieved for multiple instances of compliant trajectories that track the reference trajectory according to a Cartesian impedance law with an additional position offset. Associating a randomized position offset with each scenario achieves robust behavior. This allows one to execute the reference trajectory in conjunction with impedance-based tracking on the real system and as long as the model-reality mismatch is contained in the previously modeled perturbations reliable execution is expected. In Section \ref{sec:ocp:implaw}, the utilized Cartesian impedance law is formulated. The multi-scenario optimal control formulation is derived in Section \ref{sec:ocp:ocp}. In Section \ref{sec:ocp:cost}, we state the utilized cost function that supports determination of a common reference trajectory that results in successful task completion for all considered scenarios. Finally, in Section \ref{sec:ocp:homotopy} the homotopy procedure is detailed that sequentially tightens the smooth approximations of the collision detection and contact resolution problem to obtain solutions that closely adhere to the original nonsmooth contact dynamics.

\subsection{Impedance-based Feedback Controller} \label{sec:ocp:implaw}
Cartesian impedance control is a widely used methodology to achieve compliant behavior of robots with their environment. The considered OCP formulation optimizes for a reference trajectory $x_{\mathrm{r}} = (q_{\mathrm{r}},v_{\mathrm{r}})\in \mathbb{R}^{13}$ that is tracked by the real system $x_{\mathrm{c}} = (q_{\mathrm{c}},v_{\mathrm{c}})\in \mathbb{R}^{13}$ in a compliant manner. To formulate the impedance dynamics we follow the derivations of \cite{caccavale2008} that introduce the concept of geometrically consistent stiffness. We make the following restrictions that simplify the impedance equations. We assume so-called isotropic stiffness both in translational as in rotational space, meaning that the translational stiffness is given by $K_\mathrm{t} = k_{\mathrm{t}} I_3$, with some stiffness parameter $k_{\mathrm{t}}>0$ and the rotational stiffness is given by $K_\mathrm{r} = k_{\mathrm{r}} I_3$ with $k_{\mathrm{r}}>0$. The impedance law then reads as
\begin{equation} \label{impedancewrench}
J(x_{\mathrm{r}},x_{\mathrm{c}}) = \begin{pmatrix}
K_\mathrm{t} (\rho_{\mathrm{r}} - \rho_{\mathrm{c}}) + D_\mathrm{t}(\nu_{\mathrm{r}} - \nu_{\mathrm{c}}) \\
 \Delta (\xi_{\mathrm{r}}, \xi_{\mathrm{c}}) + D_\mathrm{r}(\omega_{\mathrm{r}} - \omega_{\mathrm{c}})
\end{pmatrix},
\end{equation}
where $J(x_{\mathrm{r}},x_{\mathrm{c}}) \in \mathbb{R}^{6}$ is the resulting force wrench by which the compliant system has to be actuated such that is retains the reference state. We have to account for the dimension difference of quaternions and their corresponding velocity space. To this end, first the quaternion difference between reference and compliant orientation is considered and calculated by
\begin{align*}
(\eta_{\mathrm{cr}}, \varepsilon_{\mathrm{cr}}) &= \xi_{\mathrm{c}}^{-1} \otimes \xi_{\mathrm{r}},
\end{align*}
where $\xi_{\mathrm{c}}^{-1}$ is the inverse quaternion and $\eta_{\mathrm{cr}} \in \mathbb{R}$ is the scalar part and $\varepsilon_{\mathrm{cr}} \in \mathbb{R}^{3}$ is the imaginary part of this quaternion product. The quaternion mismatch is then accounted in the impedance law by
\begin{align*}
\Delta (\xi_{\mathrm{r}}, \xi_{\mathrm{c}}) &= 2(\eta_{\mathrm{cr}}I_{3} + [\varepsilon_{\mathrm{cr}}]_{\times}) K_\mathrm{r} \varepsilon_{\mathrm{cr}} \\
&= 2 \eta_{\mathrm{cr}} K_\mathrm{r} \varepsilon_{\mathrm{cr}},
\end{align*}
where $[\cdot]_{\times}$ is defined via the cross product such that for $a\in \mathbb{R}^{3}, b \in \mathbb{R}^{3}$ it holds $a \times b =[a]_{\times}b$. The simplification in the second equation above holds due to the assumption of isotropic stiffness. The damping matrices $D_\mathrm{t} \in \mathbb{R}^{3\times3}$ and $D_\mathrm{r} \in \mathbb{R}^{3\times3}$ are designed such that critical damping is achieved, cf. \cite{AlbuSchaeffer2003}.

\subsection{Robustness through Ensemble Contact-Implicit Optimal Control} \label{sec:ocp:ocp}
\begin{figure}
	\centering
	\includegraphics{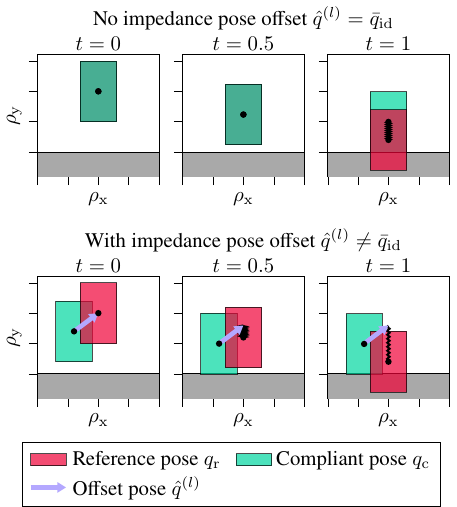}
	\caption{Illustration of impedance tracking in contact, without and with an offset pose.}
	\label{fig:impedanceillu}
\end{figure}%
In this section, we derive a robust optimal control problem formulation based on a multi-scenario approach which utilizes the derived contact dynamics of Section \ref{sec:diffphys}. To achieve robust executions on the real systems, we follow the idea of \cite{Mordatch2015}, that is optimizing for a reference trajectory that is tracked by several instances of compliant trajectories through an impedance law at a constant positional offset. In the following we introduce perturbation operators that enable the generation of these tracking offsets. We define the perturbation operator for the poses $q = (\rho, \xi) \in \mathbb{R}^{7}$ and $\hat{q}= (\hat{\rho}, \hat{\xi}) \in \mathbb{R}^{7}$ by
\begin{equation*}
P_{\mathrm{q}}(q, \hat{q}) = \begin{pmatrix}
\rho + R(\xi) \hat{\rho} \\
\xi \otimes \hat{\xi}
\end{pmatrix}\in \mathbb{R}^{7}.
\end{equation*}
The corresponding inverse perturbation operator is then given by
\begin{equation*}
	P_{\mathrm{q}}^{-1}(q, \hat{q}) = \begin{pmatrix}
		\rho - R(\xi \otimes \hat{\xi}^{-1}) \hat{\rho} \\
		\xi \otimes \hat{\xi}^{-1}
	\end{pmatrix}\in \mathbb{R}^{7},
\end{equation*}
such that it holds
\begin{equation*}
P_{\mathrm{q}}(P_{\mathrm{q}}^{-1}(q, \hat{q}), \hat{q}) = q.
\end{equation*}
We now generalize the notation such that these perturbation operators can be applied to a state $x = (q,v) \in \mathbb{R}^{13}$ as
\begin{equation*}
	P_{\mathrm{x}}(x, \hat{q}) = \begin{pmatrix}
		P_{\mathrm{q}}(q, \hat{q}) \\
		v
	\end{pmatrix}\in \mathbb{R}^{13},
\end{equation*}
and equivalently
\begin{equation*}
	P_{\mathrm{x}}^{-1}(x, \hat{q}) = \begin{pmatrix}
		P^{-1}_{\mathrm{q}}(q, \hat{q}) \\
		v
	\end{pmatrix}\in \mathbb{R}^{13}.
\end{equation*}

The perturbation operator is used to perturb the actual positions of the compliant trajectory within the impedance law. Consequently, the compliant trajectories track the reference trajectory at a constant positional offset. Figure \ref{fig:impedanceillu} illustrates this behavior, instead of tracking the reference precisely in free space the pose, obtained by adding the offset pose to the compliant pose, tracks the reference pose. If contact is established a virtual spring force is generated and applied to the compliant object by the impedance law which is opposed by the contact forces. For multiple perturbed compliant poses that track the reference pose a particle cloud originates, cf. Figure \ref{fig:title} (bottom left), consisting of the different instances of compliant trajectories that track the reference trajectory at constant positional offset in free space motions. To successfully solve an assembly problem for all instances of compliant trajectories, the optimizer has to select the reference trajectory such that contact forces are utilized to compress the particle cloud in a way such that the goal position is obtained for all scenario instances. To this end, we consider $n_{\mathrm{s}} \in \mathbb{N}$ scenarios with corresponding positional offsets $\hat{q}^{(l)} \in \mathbb{R}^{7}$, $l = 1,\dots, \nns$.

The positional offsets may be chosen at random, but since increasing the number of scenarios heavily influences computation times, selecting the offsets purposefully by hand enables more robust solutions with less computation effort. We only use translational offsets, i.e., 
\begin{equation*}
\hat{q}^{(l)} = (\hat{\rho}^{(l)},\bar{\xi}_{\mathrm{id}}),
\end{equation*}
where the specific selection of $\hat{\rho}^{(l)} \in \mathbb{R}^{3}$ is detailed for each conducted experiment in Section \ref{sec:results}. In particular, we define the translational offsets $\hat{\rho}^{(l)}$ via a normalized direction vector $\hat{\rho}^{(l)}_{\mathrm{dir}} \in \mathbb{R}^{3}$ with $\Vert \hat{\rho}^{(l)}_{\mathrm{dir}} \Vert_{2} = 1$ and an offset magnitude $\hat{\delta} \geq 0$ such that we have
\begin{equation*}
\hat{\rho}^{(l)} = \hat{\delta}\,\hat{\rho}^{(l)}_{\mathrm{dir}}.
\end{equation*}
Therefore, $\hat{\delta}$ indicates how robust a solution of the optimal control problem is but also how difficult it is to solve.

For a given initial state $\bar{x}_{0}\in \mathbb{R}^{13}$, a goal state $\bar{x}_{\mathrm{goal}}\in \mathbb{R}^{13}$ and an optimization horizon $N \in \mathbb{N}$, the multi-scenario-based optimal control formulation is now given by
\begin{mini}
	{\bm{\mathrm{x}}, \bm{\mathrm{\lambda}}_{\mathrm{n}}}{C(\bm{\mathrm{x}},\bar{x}_{\mathrm{goal}})\phantom{aaaaaaaaaaaaaaaaaaaa}} {\label{discretizedOCP}}{}
	\addConstraint{x_{\mathrm{r},0} = \bar{x}_{0},}
	\addConstraint{q_{\mathrm{r},k+1} = q_{\mathrm{r},k} + \Delta tQ(q_{\mathrm{r},k})v_{\mathrm{r},k+1},}{}
	\addConstraint{\text{for } k = 0, \dots, N-1,}
	\addConstraint{x_{\mathrm{c},0}^{(l)}= 	P_{\mathrm{x}}^{-1}(\bar{x}_{0}, \hat{q}^{(l)}),}{}
	\addConstraint{H_{\sigma,\tau}(\xrkl,\xrkpl,U_{k}^{(l)},\lambda_{\mathrm{n},k}^{(l)}) = \bm{0},}{}
	\addConstraint{G_{\sigma,\tau}(\xrkl,\xrkpl,U_{k}^{(l)},\lambda_{\mathrm{n},k}^{(l)}) \leq \bm{0},}{}
	\addConstraint{U_{k}^{(l)}}{ = J(x_{\mathrm{r},k+1},	P_{\mathrm{x}}(\xrkpl, \hat{q}^{(l)}))}
	\addConstraint{\text{for } k = 0, \dots, N-1, \ l = 1,\dots,\nns, \ }
\end{mini}
where $\bm{\mathrm{x}} = (x_{\mathrm{c},0}^{(1)},\dots, x_{\mathrm{c},N}^{(1)},\dots,  x_{\mathrm{c},N}^{(\nns)},x_{\mathrm{r},0},\dots, x_{\mathrm{r},N})$ captures all occurring states of compliant and reference trajectories and $\bm{\mathrm{\lambda}}_{\mathrm{n}} = (\lambda_{\mathrm{n},0}^{(1)},\dots, \lambda_{\mathrm{n},N}^{(1)},\dots, \lambda_{\mathrm{n},N}^{(\nns)})$ captures all occurring contact forces. The effective control input are the reference velocities $v_{\mathrm{r},k+1}, k = 0,\dots, N-1$, which uniquely define all other variables in the case of using Scholtes' smoothing. If Scholtes' relaxation is used, additional degrees of freedom are introduced due to the relaxed force-distance complementarity conditions which can be interpreted as additional control inputs that the optimizer utilizes to optimize the cost function.

\subsection{Cost Function} \label{sec:ocp:cost}
In the following, we discuss cost terms used to produce robust assembly motions. The considered cost function contains several differently weighted terms that aim at facilitating motions that can be reliably executed on real hardware. We first introduce two running cost terms
\begin{align*}
	\bar{L}_{\mathrm{run},1}(x) &= \Vert \nu \Vert_{2}^{2}, \\
	\bar{L}_{\mathrm{run},2}(x) &= \Vert \omega \Vert_{2}^{2},
\end{align*}
which quadratically penalize translational and rotational velocities. These terms effectively penalize the geometric lengths of the resulting trajectories. We further define two terminal cost terms
\begin{align*}
	\bar{L}_{\mathrm{trm},1}(x,\bar{x}_{\mathrm{goal}}) &=\Vert \rho -\bar{\rho}_{\mathrm{goal}} \Vert_{2}^{2}, \\
	\bar{L}_{\mathrm{trm},2}(x,\bar{x}_{\mathrm{goal}}) &=\left\Vert R\left(\frac{\xi}{\Vert \xi \Vert_2}\right)- R(\bar{\xi}_{\mathrm{goal}}) \right\Vert_{\mathrm{F}}^{2},
\end{align*}
where $\Vert \cdot \Vert_{\mathrm{F}}$ denotes the Frobenius norm and, as a remainder, $R(\cdot)$ is the rotation matrix associated with a quaternion. These two cost terms penalize translational and rotational mismatch to the desired goal state. Note that all of these cost terms naturally allow to construct a Gauss-Newton Hessian approximation.

Given the various cost terms, we now define the running cost for the compliant and reference trajectories as
\begin{align*}
	L_{\mathrm{run,c}}(x) &= \beta_{\mathrm{c},1} \bar{L}_{\mathrm{run},1}(x) + \beta_{\mathrm{c},2} \bar{L}_{\mathrm{run},2}(x), \\
	L_{\mathrm{run,r}}(x) &= \beta_{\mathrm{r},1} \bar{L}_{\mathrm{run},1}(x) + \beta_{\mathrm{r},2} \bar{L}_{\mathrm{run},2}(x),
\end{align*}
with scalar weights $\beta_{\mathrm{c},1}, \beta_{\mathrm{c},2},\beta_{\mathrm{r},1}, \beta_{\mathrm{r},2} \geq 0$. For the terminal costs we have
\begin{align*}
	L_{\mathrm{trm,c}}(x,\bar{x}_{\mathrm{goal}}) &= \beta_\mathrm{c,3} \bar{L}_{\mathrm{trm},1}(x,\bar{x}_{\mathrm{goal}}) + \beta_\mathrm{c,4} \bar{L}_{\mathrm{trm},2}(x,\bar{x}_{\mathrm{goal}})\\
	L_{\mathrm{trm,r}}(x,\bar{x}_{\mathrm{goal}}) &= \beta_\mathrm{r,3} \bar{L}_{\mathrm{trm},1}(x,\bar{x}_{\mathrm{goal}}) + \beta_\mathrm{r,4} \bar{L}_{\mathrm{trm},2}(x,\bar{x}_{\mathrm{goal}}),
\end{align*}
with weights $\beta_\mathrm{c,3},\beta_\mathrm{c,4},\beta_\mathrm{r,3},\beta_\mathrm{r,4} \geq 0$. Generally, the main goal when solving the OCP \eqref{discretizedOCP} is to bring all compliant trajectories to the goal state, thus, in the experiments in \mbox{Section \ref{sec:results}}, the terminal cost terms are weighted substantially more heavily than the running cost terms. The final cost function is now obtained by
\begin{align*}
C(\bm{\mathrm{x}},\bar{x}_{\mathrm{goal}}) &= \sum_{k = 0}^{N-1} \sum_{l = 1}^{\nns} L_{\mathrm{run,c}}(\xrkpl) \\
&+ \sum_{l = 1}^{\nns} L_{\mathrm{trm,c}}(x_{\mathrm{c},N}^{(l)},\bar{x}_{\mathrm{goal}}) \\
	&+ \sum_{k = 0}^{N-1} L_{\mathrm{run,r}}(x_{\mathrm{r},k+1})\\& +  L_{\mathrm{trm,r}}(x_{\mathrm{r},N},\bar{x}_{\mathrm{goal}}).
\end{align*}

\subsection{Homotopy Procedure} \label{sec:ocp:homotopy}
To facilitate the solution process, we apply a homotopy procedure to enable first exploration and then refinement of contact-rich motions. Initially the problem is solved for large $\tau_{1} \gg 0$ and $\sigma_{1} \gg 0$, resulting in relaxed object shapes and force distance complementarities, cf. Figure \ref{fig:title}. The smoothing/relaxation parameters are then reduced by update rates $\kappa_{\tau}\in (0,1)$ and $\kappa_{\sigma}\in (0,1)$ and the problem is solved again using the previous solution as a warm-start for both primal and dual variables of \eqref{discretizedOCP}. We use \texttt{IPOPT} \cite{ipopt} as solver and additionally decrease \texttt{IPOPT}'s initial barrier parameter $\mu_{\mathrm{init},1}\gg 0$. Therefore, we consider the update rate $ \kappa_\mu \in (0,1)$. For a fixed number of homotopy iterations $n_{\mathrm{hom}} \in \mathbb{N}$, in each homotopy iteration the parameters are updated as
\begin{align*}
	\tau_{n+1} &= \kappa_{\tau} \tau_{n}, \\
	\sigma_{n+1} &= \kappa_{\sigma} \sigma_{n}, \\
	\mu_{\mathrm{init},n+1} &= \kappa_\mu \mu_{\mathrm{init},n},
\end{align*}
for $n = 1,\dots, n_{\mathrm{hom}} - 1$.

\section{Numerical Evaluation and Real-world Experiments} \label{sec:results}
In the following, we first discuss the implementation framework and give results on the computational efficiency of our SDF implementation, comparing it to the performance reported on a state-of-the-art GJK-based method. We then discuss results on several peg-in-hole tasks, where we investigate required solver iterations and solving times, and illustrate the benefit of using exact Hessians compared to Gauss-Newton or L-BFGS approximations as well as using Scholtes' relaxation compared to Scholtes' smoothing. Finally, we discuss results on real-world experiments with an UR10e robot. To this end, we examine the reliability of successful task completion in dependence of the modeled uncertainty $\hat{\delta}$ and the amount of utilized smoothing $\tau,\sigma$. A video illustrating the experiments is available here: \url{https://youtu.be/g4E83bjs7lg}.

\subsection{Implementation Framework}
The code is completely written in \texttt{C/C++} and all computations are carried out on a Lenovo ThinkPad P15 Laptop with Intel i7-11850H CPU and 32 GB RAM.  We use \texttt{IPOPT} \cite{ipopt} as optimization solver with the linear solver $\texttt{MA27}$ from the \texttt{C++} interface of \texttt{CasADi} \cite{casadi}. For contact information and corresponding derivative evaluations, we use the high-performance linear algebra library \texttt{BLASFEO} \cite{blasfeo}. It provides performance-optimized linear algebra routines for matrices up to a couple hundreds elements, as typically encountered in embedded optimization. The signed distance problem \eqref{sdfopt} is solved by using the high-performance interior-point implementation \texttt{HPIPM} \cite{hpipm}, which is based on \texttt{BLASFEO}. The nominal evaluation of the contact information vector $w$ \eqref{contactinformation} and all first- and second-order derivatives as occurring in Section \ref{sec:colldet} are analytically derived and evaluated using \texttt{BLASFEO}. Within the OCP \eqref{discretizedOCP} a total of $\vert \Pi \vert\cdot \nns \cdot N$ contact information vectors and corresponding derivatives are evaluated per solver iteration. To expose the distance information to \texttt{CasADi}, we utilize \texttt{CasADi}'s \texttt{Callback} class. To speed up the SDF evaluation procedure, we use \texttt{OpenMP} for parallel computations in combination with a batched version of the \texttt{Callback} class, which takes all $\nns \cdot N$ object positions occuring in \eqref{discretizedOCP} and returns all the corresponding contact information at once.
 
\subsection{Efficiency of SDF}

\begin{figure}
	\centering
	\includegraphics{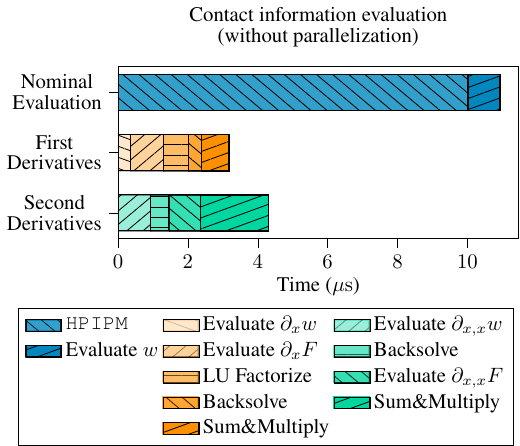}
	\caption{Computation time for a single cube-cube distance (averaged from 1000 evaluations). Here, $x \in \{q,\gamma \}$ is a placeholder for the respective derivatives detailed in \mbox{Section \ref{sec:colldet}}.}
	\label{fig:sdfresults}
\end{figure}

First, we give an analysis of computation time required for our SDF implementation to evaluate signed distances and to calculate first- and second-order directional derivatives. To this end, we evaluate 1000 cube-cube distances, where one cube has a fixed position and one cube is placed at a randomly generated position. Figure \ref{fig:sdfresults} shows the average time required for one LP solution, its Jacobian and the directional Hessian for a randomly generated seed. We can deduce that evaluating first-derivatives for this problem is approximately three times cheaper than solving the distance problem itself. Moreover, evaluating second-derivatives is only slightly more expensive than evaluating first-derivatives in terms of computation times, making derivative evaluations in this setup overall very cheap compared to the computational effort for nominal evaluations.

In \cite{Montaut2024}, the state-of-the-art QP solver \texttt{ProxQP} \cite{ProxQP} has been used to solve similiar distance problems, where a computation time of $5 \mu\mathrm{s}$ has been reported. \texttt{ProxQP} is a first-order method not relying on the inversion of the KKT system to solve QPs, making it cheap in computation. However, it does not provide an inherent smoothing mechanism as an interior-point method does. The efficiency of \texttt{HPIPM} is illustrated as we only require $10 \mu\mathrm{s}$ to solve a distance problem in a smooth manner. Arguably, state-of-the-art GJK algorithms report computation times for similar distance problems of $0.2 \mu\mathrm{s}$ \cite{Montaut2024}. This points to an interesting area of research that aims at combining the advantages of smoothing using interior-point methods with the efficiency of GJK methods.

\subsection{Peg-in-Hole Assembly with Randomized Shapes} \label{sec:results:peginhole}
\begin{figure*}[htbp]
	\centering
	
	\begin{subfigure}[b]{0.19\textwidth}
		\centering
		\includegraphics[width=\textwidth, trim=30 20 30 16, clip]{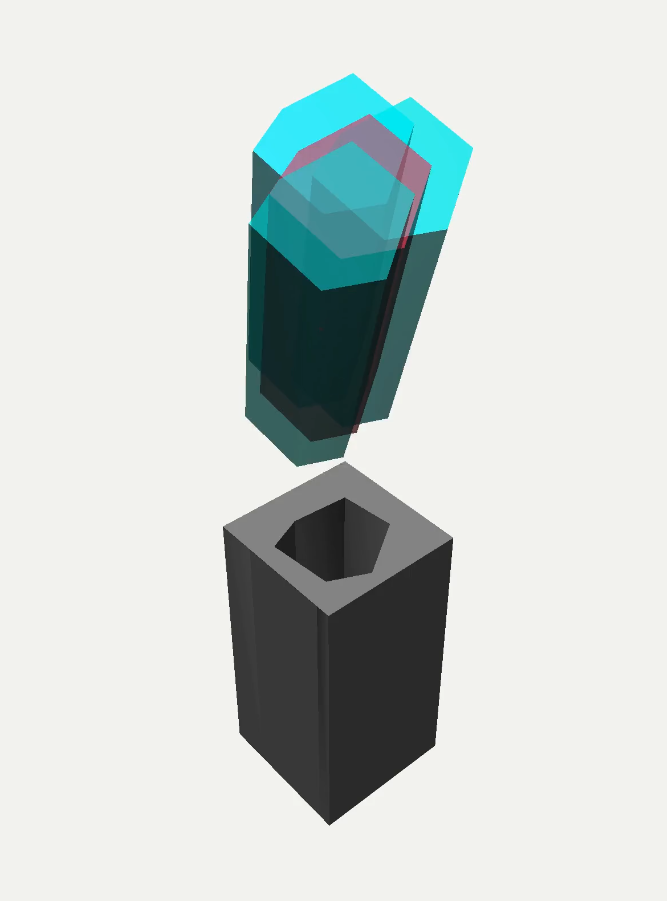}
		\caption{\texttt{Peg1}: $\vert \Pi \vert= 6$}
		\label{fig:subfig-a}
	\end{subfigure}
	\begin{subfigure}[b]{0.19\textwidth}
		\centering
		\includegraphics[width=\textwidth, trim=30 20 30 10, clip]{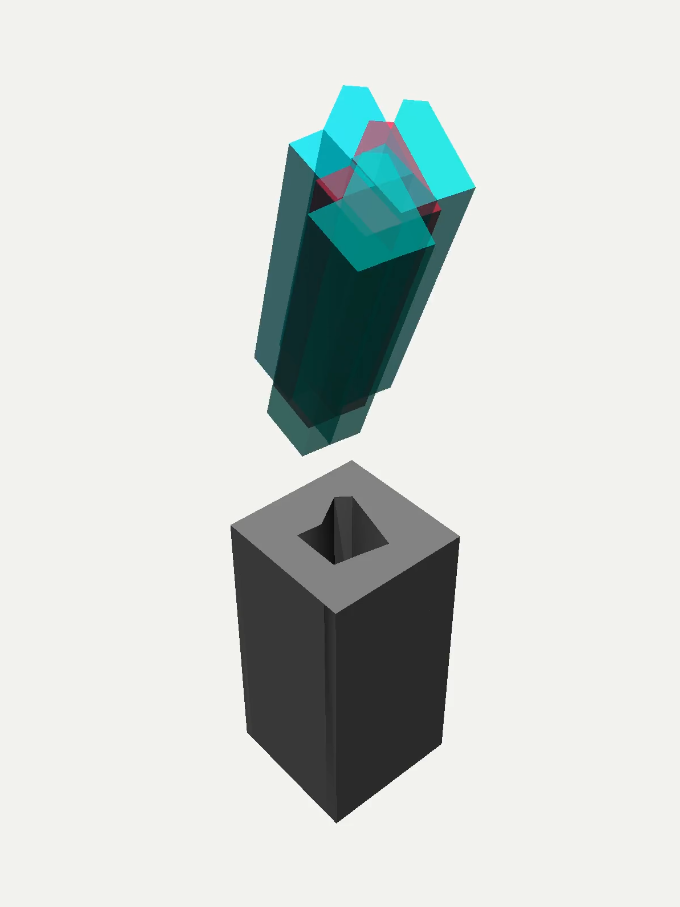}
		\caption{\texttt{Peg2}: $\vert \Pi \vert= 10$}
		\label{fig:subfig-b}
	\end{subfigure}
	\begin{subfigure}[b]{0.19\textwidth}
		\centering
		\includegraphics[width=\textwidth, trim=30 20 30 10, clip]{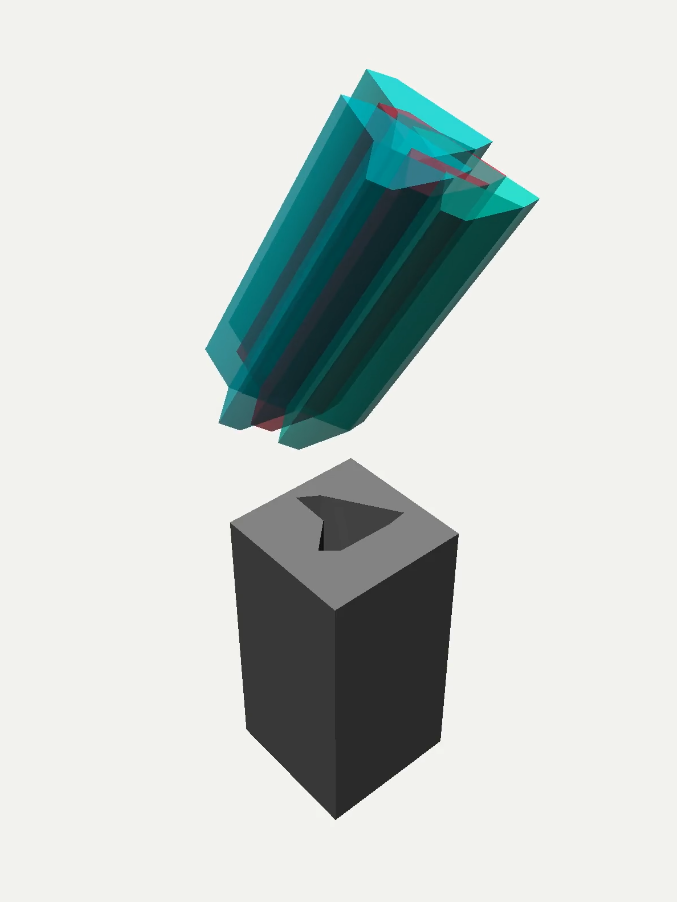}
		\caption{\texttt{Peg3}: $\vert \Pi \vert= 14$}
		\label{fig:subfig-c}
	\end{subfigure}
	\begin{subfigure}[b]{0.19\textwidth}
		\centering
		\includegraphics[width=\textwidth, trim=30 20 30 16, clip]{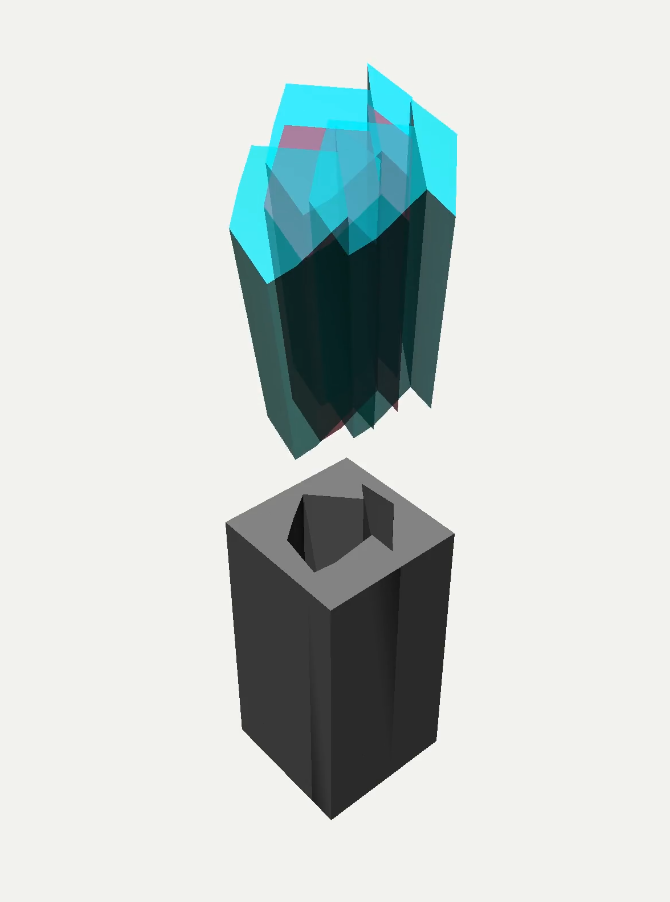}
		\caption{\texttt{Peg4}: $\vert \Pi \vert= 18$}
		\label{fig:subfig-d}
	\end{subfigure}
	\begin{subfigure}[b]{0.19\textwidth}
		\centering
		\includegraphics[width=\textwidth, trim=30 20 30 10, clip]{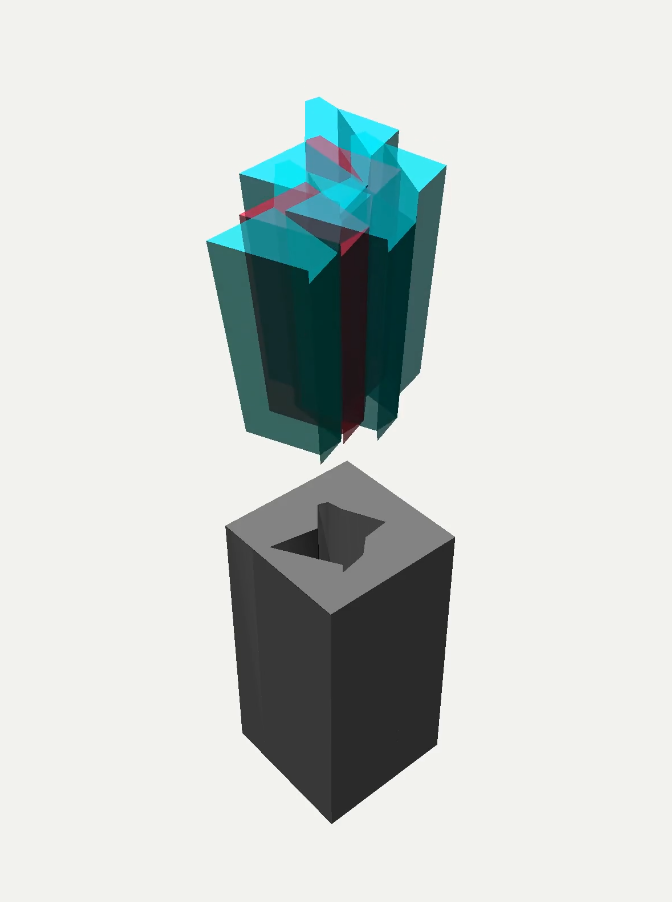}
		\caption{\texttt{Peg5}: $\vert \Pi \vert= 24$}
		\label{fig:subfig-e}
	\end{subfigure}
	
	\caption{Five different considered Peg-in-Hole problems with increasing numbers of contact pairs.}
	\label{fig:peginhole_vis}
\end{figure*}

\begin{table}
	\centering
	\caption{Average total computation time for the peg-in-hole experiments.}
	\label{table:total_comp_times}
	\vspace{-3.8pt}
	\begin{tabular*}{\columnwidth}{c | c | c | c | c | c}
		Problem instance time& \texttt{Peg1} & \texttt{Peg2} & \texttt{Peg3} & \texttt{Peg4} & \texttt{Peg5} \\
		\hline
		Total wall time ($\mathrm{s}$) & 85.76 & 171.33 & 391.22 & 380.18 & 343.57 
	\end{tabular*}
\end{table}

\begin{figure}
	\centering
	
	\begin{subfigure}{0.4\textwidth}
		\centering
		\includegraphics{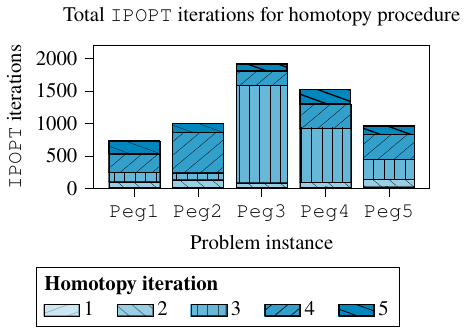}
		\vspace{-1em}
		\caption{Average number of \texttt{IPOPT} iterations during each homotopy iteration.}
		\label{fig:pegs_data:iterations}
	\end{subfigure}
	\begin{subfigure}{0.4\textwidth}
		\centering
		\vspace{1em}
		\includegraphics{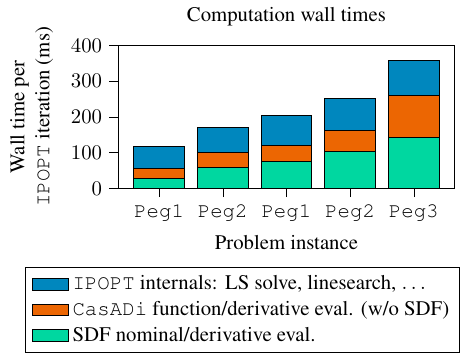}
		\caption{Average wall time per \texttt{IPOPT} iteration splitted into time required for SDF, contact normal and corresponding derivative evaluations, \texttt{CasADi} evaluations for terms in the OCP \eqref{discretizedOCP} unrelated to the SDF, and remaining time spent by \texttt{IPOPT} for computing the next iterate.}
		\label{fig:pegs_data:times}
	\end{subfigure}
	
	\caption{Each peg-in-hole problem is solved for five starting positions with different orientations. The figures show the average results across these five solves. Parameters used for these experiments are denoted in Table \ref{table:parameters_pegs} in Appendix \ref{appendix:B}.}
	\label{fig:pegs_data}
\end{figure}

We analyze the performance of the considered methodology on five different peg-in-hole problems as shown in Figure \ref{fig:peginhole_vis}. The shapes of the pegs have been randomly generated, where the number of resulting contact pairs ranges from $\vert \Pi \vert = 6$ to $\vert \Pi \vert = 24$. Each peg-in-hole problem is solved for five different initial poses, where the translational position was kept equal and only the orientation was altered. Table \ref{table:parameters_pegs} in Appendix \ref{appendix:B} denotes all parameters that were used to solve the corresponding instances of the OCP \eqref{discretizedOCP}. 

Figure \ref{fig:pegs_data} shows the results of this experiment. We observe in Figure \ref{fig:pegs_data:iterations}  that the geometric complexity of the pegs shapes is not directly proportional to the required iterations to achieve convergence. This is reasonable as robust assembly behavior for pegs is achieved by tilting the peg and consecutively pushing it into one corner of the hole, whether there are a few or many more other corners does not directly influence the complexity of the motion (illustrated in the accompanying video).

In contrast, the required computation times per iteration as shown in Figure \ref{fig:pegs_data:times} are directly proportional to the complexity of the peg's shapes since more considered contact pairs increase the number of contact information evaluations as well as the number of optimization variables and constraints in the OCP \eqref{discretizedOCP}. This directly influences $\texttt{IPOPT}$, $\texttt{CasADi}$ and SDF computation times. \texttt{IPOPT} is currently used with the linear solver \texttt{MA27} which does exploit generic sparsity but is not tailored to the specific sparsity implied by the optimal control problem. Thus, there is room for improvement by using sparsity exploiting linear solvers tailored to optimal control as, e.g., done in \cite{Fatrop}. Nevertheless, the proposed method is already highly efficient, as can be deduced from Table \ref{table:total_comp_times}, where we demonstrate that robust assembly motions can be determined for the considered peg-in-hole instances in a matter of one to seven minutes on a standard laptop CPU.

In \cite{Vuong2023}, a gradient-free reinforcement learning approach in conjunction with the \texttt{MuJoCo} physics engine is utilized to solve peg-in-hole problems, requiring a number of simulation steps in the order of $10^{8}$ for successful task completion. This also means that all the collision detection problems required for a simulation step must be solved as many times as the total number of simulation steps. With our optimal control approach, we require an order of $10^{3}$ iterations to converge to a solution, where in each iteration $N = 10^{2}$ simulation steps are approximately evaluated by calculating one Newton step on the dynamics, causing overall an order of $10^{5}$ calls to the collision detection pipeline. This supports the potential of increased sample-efficiency that optimal control approaches inherit compared to methods that use less derivative information of the dynamic model.

\begin{figure}
	\centering
	\includegraphics{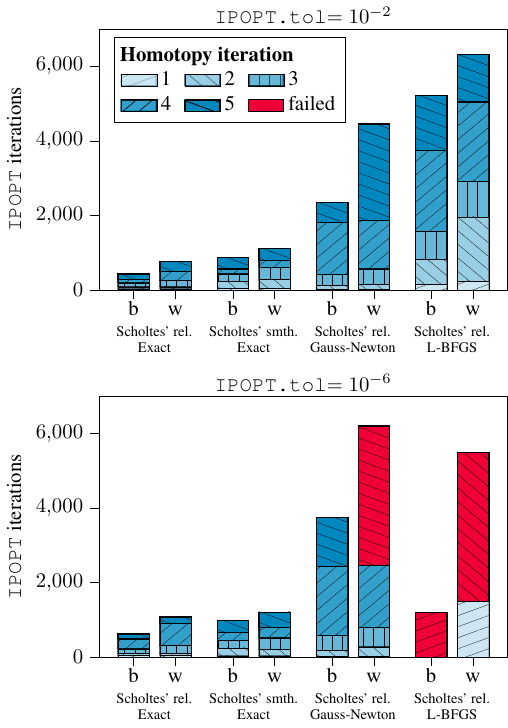}
	\caption{Comparison of Scholtes' relaxation (rel.) vs Scholtes' smoothing (smth.) and utilization of exact, Gauss-Newton and L-BFGS Hessians for the solution of the OCP \eqref{discretizedOCP}. The \texttt{Peg1} problem is solved for five different starting positions for each of the considered instances, where the best (b) and the worst (w) trial in terms of solver iterations is shown. The experiments are repeated for two different values of the convergence tolerance \texttt{IPOPT.tol}.}
	\label{fig:hessianapprox_comp}
\end{figure}

We now also investigate whether using exact second-order derivatives over commonly used Hessian approximations is beneficial in the presented optimal control approach. Further, we compare computational performance achieved with Scholtes' relaxation vs. Scholtes' smoothing. To this end, Figure \ref{fig:hessianapprox_comp} shows the results for solving the \texttt{Peg1} problem with Scholtes' relaxation and Scholtes' smoothing as well as with exact, Gauss-Newton or L-BFGS Hessians. The best and worst run in terms of iteration numbers out of five problem instances with different initial peg poses is shown for two different values of \texttt{IPOPT}'s convergence tolerance parameter \texttt{IPOPT.tol}. We deduce that Scholtes' relaxation improves convergence speed compared to using Scholtes' smoothing. This illustrates that derivative-based planning and control approaches using differentiable physic simulations can strongly benefit if they can influence additional degrees of freedom in the physics simulation. \\
Furthermore, these results emphasize clearly the benefit of using exact Hessians compared to using a Gauss-Newton or L-BFGS approximation for the conducted experiments. In case of the loose convergence tolerance \texttt{IPOPT.tol} $= 10^{-2}$ exact Hessians result in 5 times faster convergence than for the Gauss-Newton Hessian and 10 times faster convergence than for the L-BFGS Hessian. In case of \texttt{IPOPT}'s default convergence tolerance \texttt{IPOPT.tol} $= 10^{-6}$, both Hessian approximations lead to failed convergence, where for Gauss-Newton three out of five optimization attempts fail while for L-BFGS all optimization attempts fail.

\subsection{Clamp Assembly} \label{sec:results:clamp}
\begin{figure*}[ht]
	\centering
	\includegraphics[width=\textwidth]{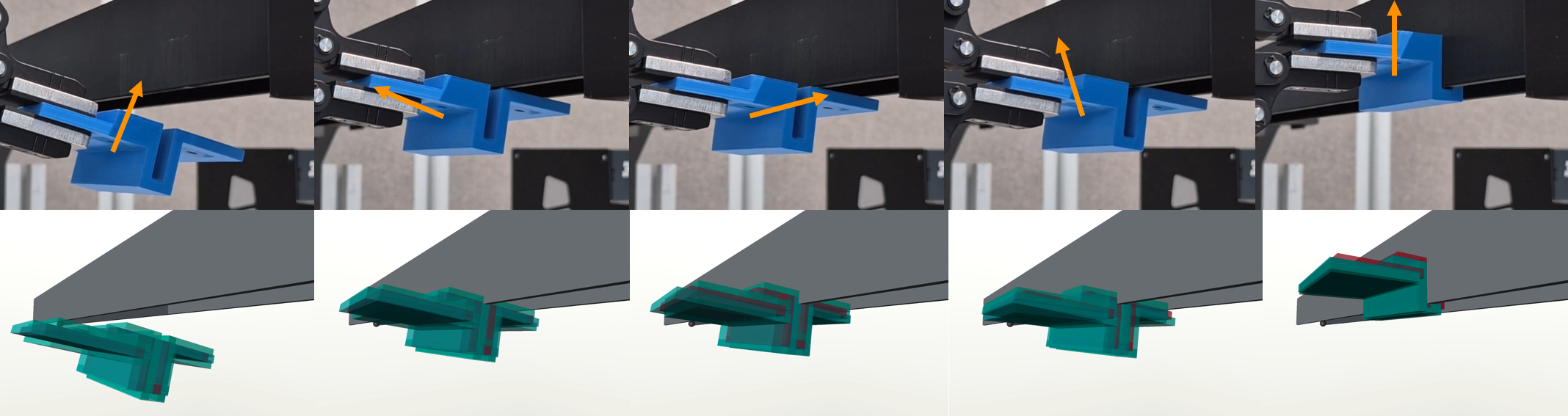}
	\caption{Solution trajectory for the considered benchmark problem. Orange arrow indicates the current moving direction. Best understood by viewing the accompanying video.}
	\label{fig:clampassemblyvis}
\end{figure*}
\begin{figure}[htbp]
	\centering
	\begin{subfigure}[b]{0.48\linewidth}
		\centering
		\includegraphics[width=\linewidth]{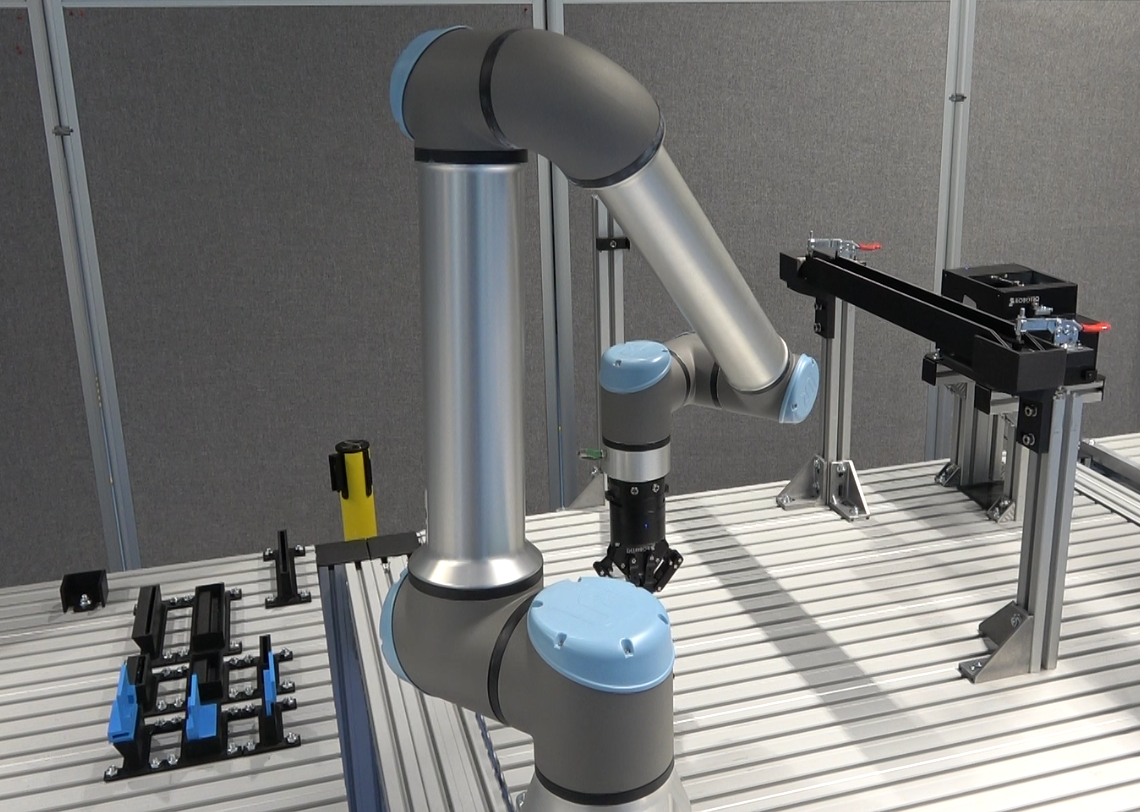}
		\caption{Overview of robotic setup.}
		\label{fig:1a}
	\end{subfigure}
	\hfill
	\begin{subfigure}[b]{0.48\linewidth}
		\centering
		\includegraphics[width=\linewidth,trim=0 80 0 80, clip]{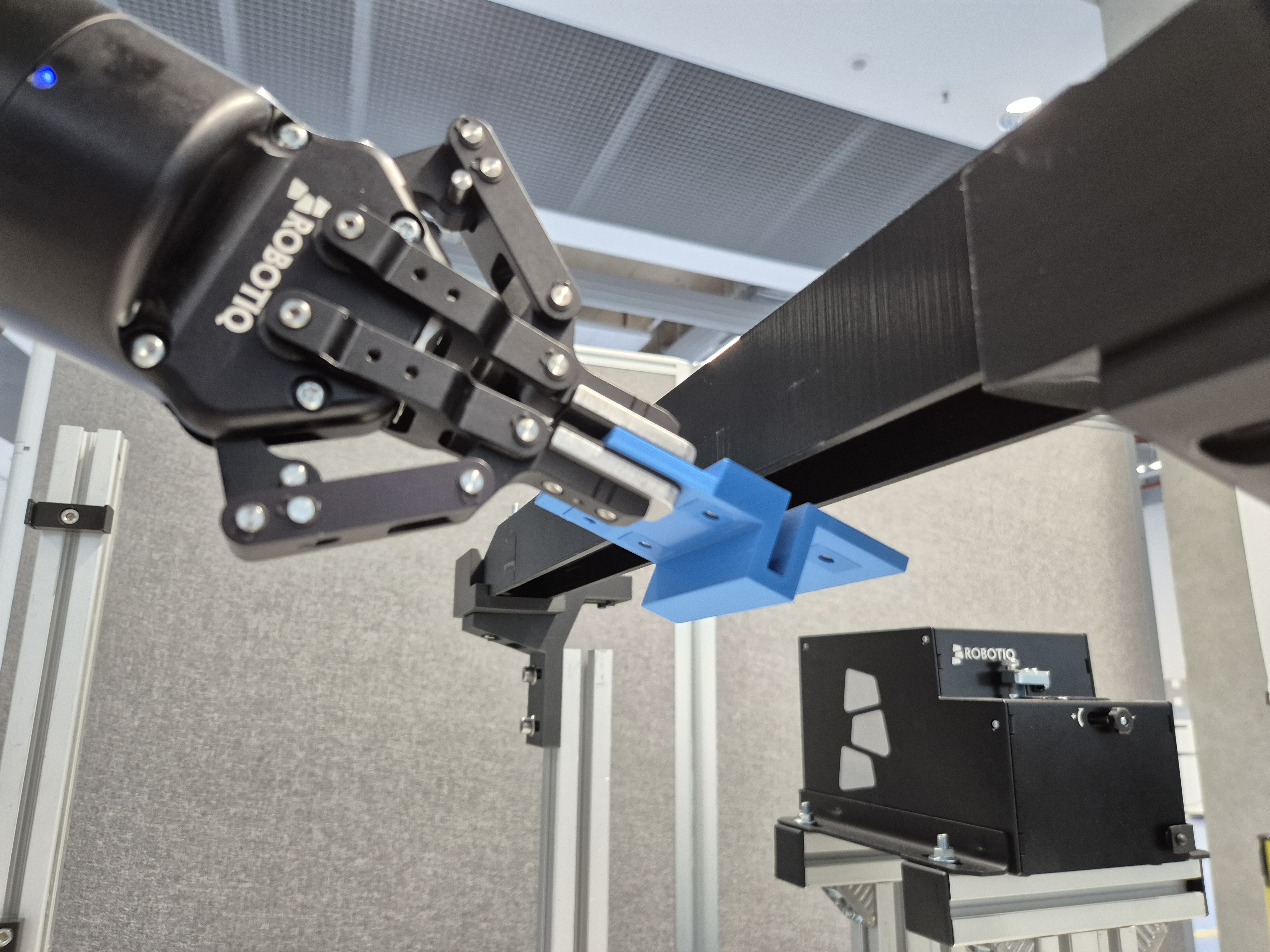}
		\caption{\texttt{Clamp1} assembly.}
		\label{fig:1b}
	\end{subfigure}
	
	\vspace{4mm} 
	
	\begin{subfigure}[b]{0.48\linewidth}
		\centering
		\includegraphics[width=\linewidth,trim=0 80 0 80, clip]{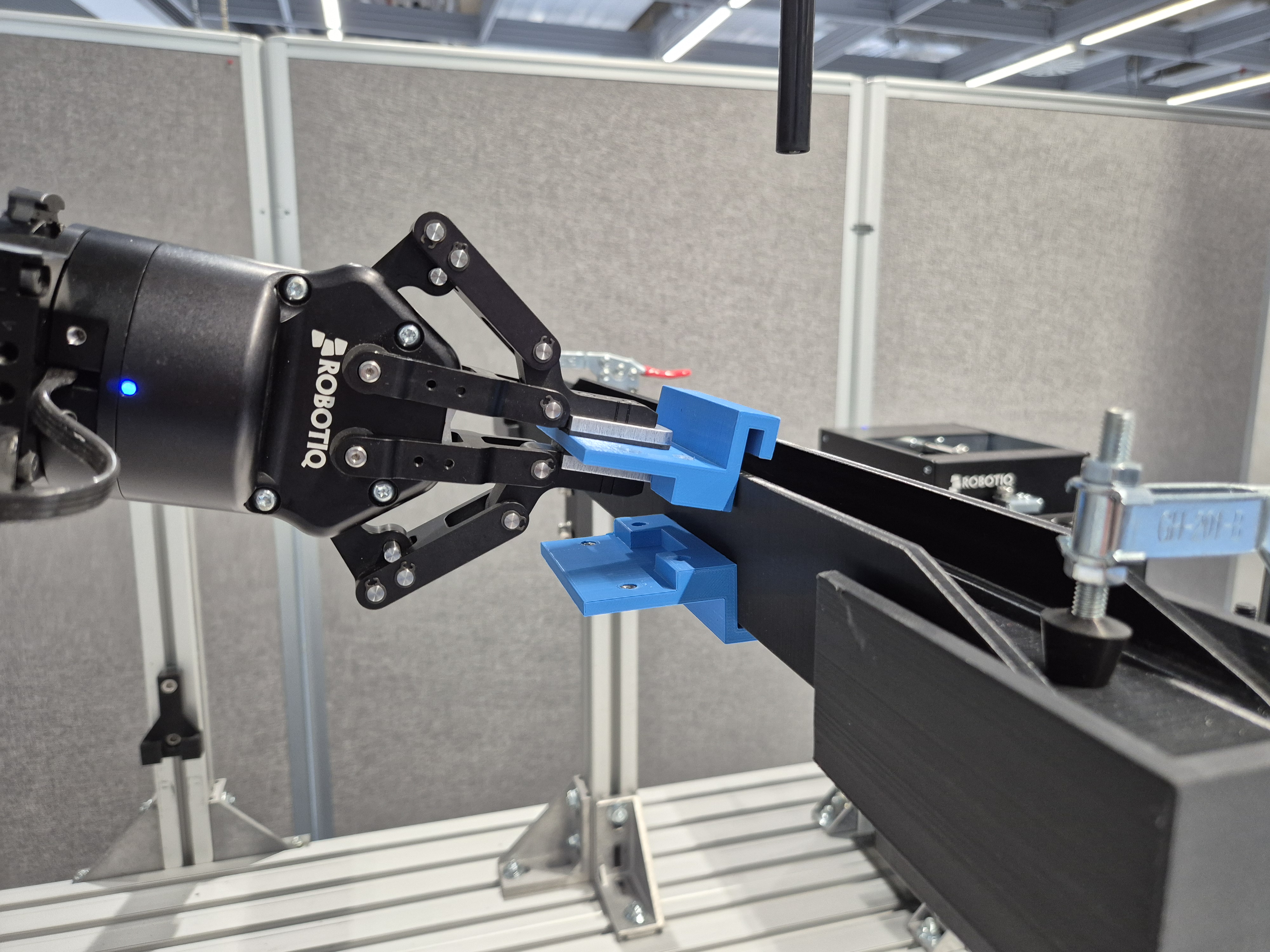}
		\caption{\texttt{Clamp2} assembly.}
		\label{fig:2a}
	\end{subfigure}
	\hfill
	\begin{subfigure}[b]{0.48\linewidth}
		\centering
		\includegraphics[width=\linewidth,trim=0 80 0 80, clip]{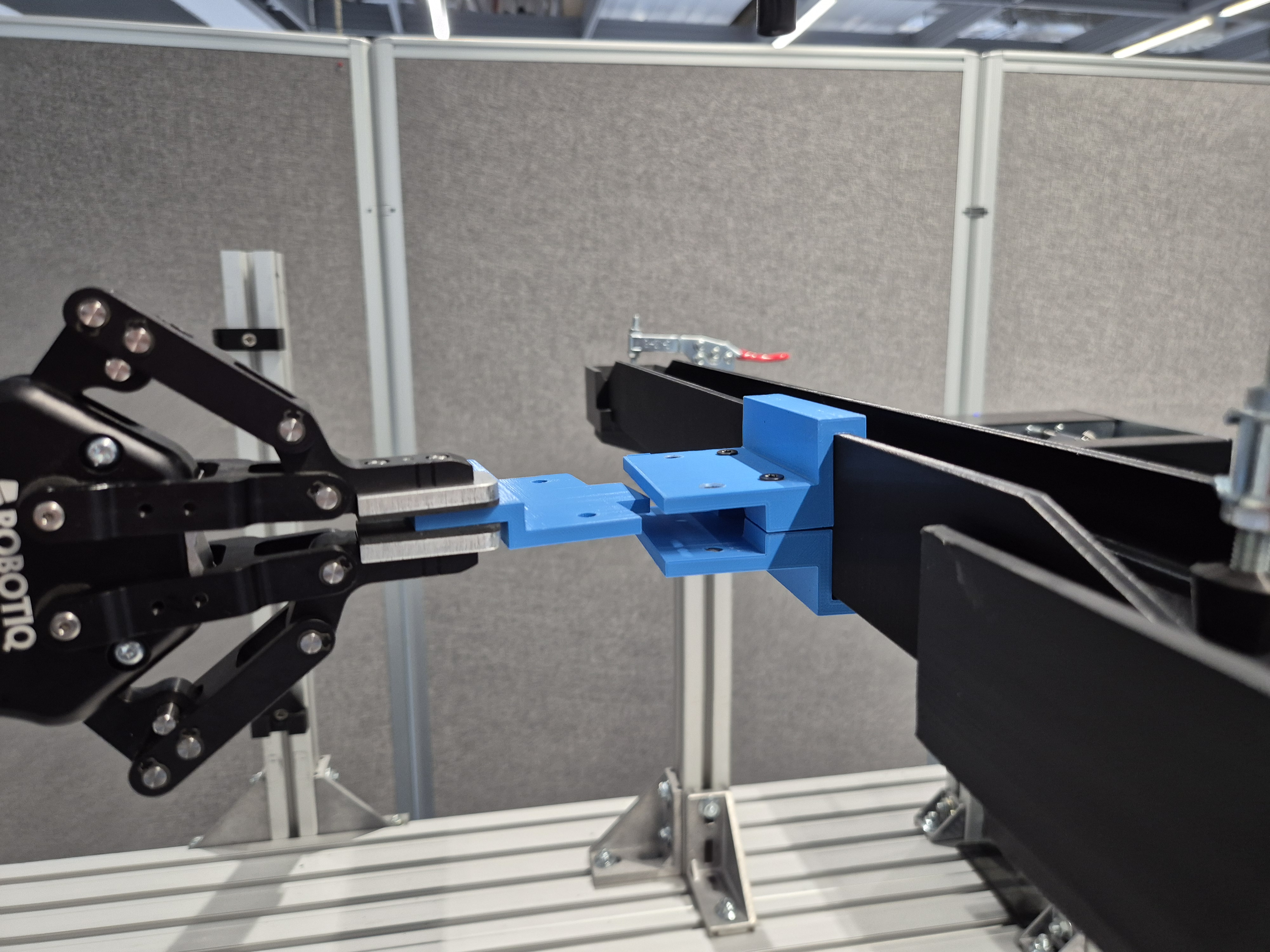}
		\caption{\texttt{Clamp3} assembly.}
		\label{fig:2b}
	\end{subfigure}
	
	\caption{The robotic setup for the conducted experiments. An UR10e is used in conjunction with the proposed methodology to robustly assemble three different parts onto an H-strut.}
	\label{fig:roboticsetup}
\end{figure}
Next, we conduct experiments that test the reliability of the calculated robust assembly motions on real robotic hardware, cf. Figure \ref{fig:roboticsetup} for the robotic setup. In particular, we investigate what influence the smoothing parameters $\tau$ and $\sigma$ used for the final homotopy iteration and the magnitude of the modeled uncertainty $\hat{\delta}$ have on the task success rates. To this end, an UR10e manipulator is controlled via the \texttt{ur\_rtde} interface \cite{Lindvig_rtde}. We control the robot via the \texttt{forceMode}, which allows one to command a six-dimensional Cartesian wrench that the end-effector tries to exhibit on its environment. The commanded wrench is exactly the one computed by the impedance law \eqref{impedancewrench}. However, since on the real system this force wrench does not only accelerate the part but also the whole robot, we increase the stiffness values to $k_{\mathrm{t},\mathrm{robot}} = 1500$ and $k_{\mathrm{r},\mathrm{robot}}=200$ which differ from the values used in simulation $k_{\mathrm{t},\mathrm{sim}} = 100$ and $k_{\mathrm{r},\mathrm{sim}}=10$ to account for this mismatch.

We consider the task of assembling a clamp onto a H-formed strut, cf. Figure \ref{fig:clampassemblyvis}. The wall thickness of the strut is 5mm while the gap width of the clamp is 5.5mm, whereby due to 3D printing imprecision the final clearance between clamp and strut is less than 0.5mm. We carefully measured the relative position of the robot and the H-formed strut in the real-world to reduce the unknown positional deviations between the model and reality as much as possible. 

The assembly problem is then solved 30 times, where 5 different starting positions and 6 different magnitudes $\hat{\delta}$ for the positional offsets $\hat{q}^{(l)}$ are considered. Every problem is solved with up to $n_{\mathrm{hom}}=6$ homotopy iterations, resulting in $30\times6 = 180$ distinct reference trajectories. Table \ref{table:parameters_clamp} in Appendix \ref{appendix:B} denotes the utilized parameters for these experiments.

These reference trajectories are executed on the real robot by tracking them with the Cartesian impedance controller. To verify the robustness, we inject virtual model-reality mismatches by changing the position of the part within the gripper in the virtual model. This injected mismatch is precisely what is replicated by the robust OCP formulation \eqref{discretizedOCP}, as now the part in the real system does not directly track the reference trajectory, but instead tracks it at a constant positional offset. We inject two different positional offsets, one in distribution $q^{\mathrm{inject},(1)}$, meaning there is a $\hat{q}^{(l)}$ equivalent to $q^{\mathrm{inject},(1)}$, and one out of distribution $q^{\mathrm{inject},(2)}$, on the real system.
\begin{figure}
	\centering
	\includegraphics{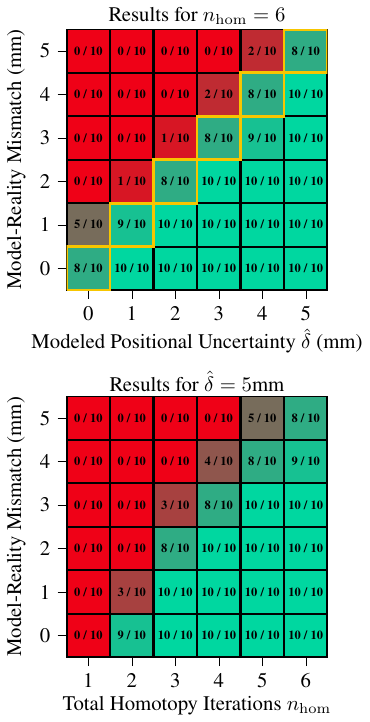}
	\caption{Results of real-world experiments with an UR10e. Robust reference trajectories where calculated for different magnitudes of modeled uncertainty $\hat{\delta}$ and final smoothing parameters $\tau,\sigma$. The figure shows the amount of successful completions of the assembly tasks. A run was termed successful if the final position error to the goal position was sufficiently small. See the videos for more details on the experiments. We observe a success rate of 80-90 \% if the model-reality mismatch matches the modeled positional uncertainty (orange highlights), and an over $99\%$ success rate if the modeled positional uncertainty is at least 1mm larger than the actual model-reality mismatch (lower right triangular submatrix of top plot).}
	\label{fig:realworldresults}
\end{figure}

Figure \ref{fig:realworldresults} shows the success rates of tested assembly motions across 720 executed motions. The top figure shows the success rates for different combinations of values of modeled positional uncertainty $\hat{\delta}$ and injected model-reality mismatches. For each combination the trajectories for the five different starting positions in conjunction with the two considered injected positional offsets are excuted, making up 10 trials per combination. If there would be no sim-to-real gap at all, one would expect that the main diagonal of the plot admits to a 100\% success rate. Since contact dynamics and the impedance controller behave differently on the real system than in simulation, less success rate is expected, in our experiments we achieve 80-90\%. 

However, the more important result is that the success rates are almost constant along the main and subdiagonals. The average result on these diagonals may be easily shifted to the left by making the assembly problem easier, e.g., through introducing larger clearances, or to the right by making the problem harder, e.g., generating more mismatch between simulated and real-world controller. Since the success rates stay constant along the diagonals, this indeed supports the hypothesis that increased model-reality mismatches can be counteracted by modeling more positional uncertainty. On top, if one estimates an upper bound for the model-reality mismatch present in ones system, one can choose the modeled positional uncertainty with large enough margin to guarantee an almost 100\% success rate. For example, in our case, by choosing the modeled uncertainty such that it is at least 1mm larger than the actual model-reality mismatches guarantees successful completion of the task in 149 out of 150 trials (this refers to the lower-right triangular submatrix).

The bottom figure now investigates what influence the final utilized smoothing parameters have on the real-world success rates. The magnitude of smoothing is naturally directly proportional to the number of utilized homotopy iterations. The solutions are calculated for a fixed modeled uncertainty $\hat{\delta} = 5$mm. Thus, the right most column of the top and bottom figure admit to the same motions that where executed on the robot. The results in this right column differ by the success of one motion, this difference likely being due to wear and tear of the 3D printed parts between the execution of the first 360 motions an the second 360 motions. Now, one can observe that the magnitude of the final smoothing parameters is directly inverse proportional to the success rates on the real system, meaning smaller smoothing of the physics simulation results in more reliable execution on the real-system.

This implies room for interesting trade-offs, as for example to guarantee a 100\% success rate on our system with a 2mm model reality-mismatch, we could either model 3mm uncertainty and use six homotopy iterations to obtain a solution with small final smoothing parameters, or instead model 5mm uncertainty while only using four homotopy iterations to obtain a solution with larger final smoothing of contact dynamics. I.e., the in this paper presented robust formulation for contact-rich optimal control allows one to flexibly swap robust modeling for contact simulation precision. This can also be particularly interesting for other control approaches such as MPC, which often relies on Gauss-Newton Hessians, which particularly struggled at homotopy iterations with small smoothing parameters as seen in Figure \ref{fig:hessianapprox_comp}.

\section{Conclusion}
We proposed a robust optimal control approach which determines contact-rich assembly motions that can be reliably executed on real robotics hardware. In particular, we have demonstrated the ability of second-order optimization methods to solve difficult contact-rich problems in a sample-efficient manner, as well as the advantage of using exact Hessians over commonly used Hessian approximations for the considered use cases. We also showed in real-world experiments that robust modeling enables transition of solutions for smoothed contact dynamics to real-world problems. Overall, the results illustrate that there is great potential for derivative-based optimization methods in conjuncture with contact-rich systems if collision detection, contact dynamics and corresponding control methods are carefully constructed. 

Current limitations are as follows. The size of the collision detection LP subproblems scales proportionally with the number of halfspaces required to represent the convex polytopes, and the number of constraints in the OCP scales proportional to the number of considered collision pairs. For problems with complex convex subshapes or many collision pairs this results in extensive computation times. Furthermore, state-of-the-art physic simulation approaches commonly include a Coulomb friction model which was not considered in this paper.

\section{Future Work}
In future work, we aim to transfer the proposed method to the multiple shooting case where the dynamics evaluation is hidden from the OCP. Solving for collision detection and contact resolution problems in an isolated manner allows one to only consider locally relevant constraints in the subproblems and, therefore, decrease computational cost for assembly problems with complex polytopic shapes and many collision pairs. Furthermore, a Coulomb friction model should be considered, a suitable formulation is, e.g., Anitescu's contact dynamics given by a quadratically constrained quadratic program \cite{Anitescu2006}. The benefit of simulating friction effects is likely problem dependent, nevertheless a study how these effects can be used together with robust modeling to achieve improved real-world execution reliability should be conducted. Furthermore, computation speed of the method should be improved by using OCP-structure exploiting linear system solvers within the optimization method.

\section*{Acknowledgments}
\noindent We thank Florian Braun for advice on the C++ implementation framework and the visual presentation. We further thank Daniel Larin for exchange on formulations aspect of contact-rich dynamics.
This research was supported by BMWK via 20D2123B, 03EI4057A and 03EN3054B, by DFG via Research Unit FOR 2401, project 424107692 and 525018088, and by the EU via ELO-X 953348.

{\appendices
\section{Derivation of Second-Order Directional Derivatives for an Implicit Function} \label{appendix:A}
We construct second-order directional derivatives for the primal-dual variables $\gamma$. These variables are defined as an implicit function according of the KKT system \eqref{KKTsystem}, dropping the dependence on $\tau$ we have
\begin{equation*}
F(\gamma, q) = \bm{0}.
\end{equation*}
As already noted in \eqref{firstderF}, the Jacobian of the primal-dual variables is obtained through solving the following linear system
\begin{equation} \label{app:IFT}
	\delgamma F(\gamma(q), q)	 \, \dq \gamma(q) = -\delq F(\gamma(q), q).
\end{equation}
Given the seed vector $s_{\gamma} \in \mathbb{R}^{n_{\gamma}}$, we now want to obtain the directional second-order derivatives
\begin{equation*}
	\langle  s_\gamma \, , \, \dqtwo\gamma(q) \rangle = \dq (\dq \gamma(q) \tsp s_\gamma).
\end{equation*}
To this end, we write for simplicity 
\begin{equation} \label{app:simpledef}
	\begin{split}
&A(q) = \delgamma F(\gamma(q), q), \\
&X(q) =  \dq \gamma(q), \\
&B(q) = -\delq F(\gamma(q), q).
	\end{split}
\end{equation}
Then \eqref{app:IFT} takes the form of the block linear system
\begin{equation}
A(q) X(q) = B(q).
\end{equation}
Differentiating with respect to a scalar component of the pose $q_k$, we obtain by the chain rule
\begin{align*}
	&\mathrm{D}_{q_{k}} A(q) X(q) +  A(q)\mathrm{D}_{q_{k}} X(q) =\mathrm{D}_{q_{k}} B(q) \\
	\Leftrightarrow \ \ & \mathrm{D}_{q_{k}} X(q) =  A(q)^{-1}(\mathrm{D}_{q_{k}} B(q) - \mathrm{D}_{q_{k}} A(q) X(q)) \\
	\Rightarrow \ \ &  s_\gamma \tsp \mathrm{D}_{q_{k}} X(q) = s_\gamma \tsp A(q)^{-1}(\mathrm{D}_{q_{k}} B(q) - \mathrm{D}_{q_{k}} A(q) X(q)).
\end{align*}
Thus, the adjoint equation \eqref{secondderadjeq} originates given by
\begin{equation*}
	 A(q)^{T} r_\gamma = s_\gamma.
\end{equation*}
We can now substitute $r_\gamma$ and transpose the equations to obtain column vectors
\begin{align*}
	\mathrm{D}_{q_{k}}  X(q) \tsp s_\gamma = \mathrm{D}_{q_{k}} B(q) \tsp r_\gamma - X(q)\tsp \mathrm{D}_{q_{k}} A(q)\tsp r_\gamma
\end{align*}
Concatenating the scalar derivatives w.r.t. $q_k$ lets one obtain the full directional Hessian
\begin{equation} \label{app:simpleform}
	\mathrm{D}_{q}  X(q) \tsp s_\gamma = \mathrm{D}_{q} B(q) \tsp r_\gamma - X(q)\tsp \mathrm{D}_{q} A(q)\tsp r_\gamma.
\end{equation}
Substituting \eqref{app:simpledef} into \eqref{app:simpleform} and  evaluation of the right-hand side total derivatives results in the expression \eqref{secondderfF} used in Section \ref{sec:colldet:derivatives}.

\section{Parameters for conducted experiments.} \label{appendix:B}

\newcolumntype{Y}{>{\raggedleft\arraybackslash}X}
\begin{table}[H]
	\centering
	\caption{Parameters for the peg-in-hole experiments of Section \ref{sec:results:peginhole}.}
	\label{table:parameters_pegs}
	\vspace{-3.8pt}
\begin{tabularx}{\columnwidth}{@{}l r | l r@{}}
	\hline \hline
	\textbf{Parameter} & \textbf{Value} & \textbf{Parameter} & \textbf{Value}\\
	\hline \hline
$	N $& 100  & $\beta_{\mathrm{r},1}$ & 1\\
		$\Delta t$ & 0.04 & $\beta_{\mathrm{r},2}$ & 0.1 \\
$n_{\mathrm{s}} $& 3  & $\beta_{\mathrm{r},3}$ & 100\\
$	\vert \Pi \vert $& \phantom{fillllll}cf. Figure \ref{fig:peginhole_vis} & $\beta_{\mathrm{r},4}$ & 10\\
	
	$k_{\mathrm{t}}$ & 50& $\beta_{\mathrm{c},1}$ & 1  \\
	$k_{\mathrm{r}}$ & 5 &$\beta_{\mathrm{c},2}$ & 0.1 \\
	$n_{\mathrm{hom}}$& 5 &$\beta_{\mathrm{c},3}$ & 10000 \\
	$\tau_{1} $& 0.0025 &$\beta_{\mathrm{c},4}$ & 1000 \\
	$\sigma_{1} $& 0.00125 &$\hat{\delta}$ & 0.02 \\
	$\mu_{\mathrm{init},1}$& 1 & $\hat{\rho}^{(1)}$ & $(0.7071,\,-0.7071,\,0)$  \\
$\kappa_{\tau}$ & 0.5 & $\hat{\rho}^{(2)}$ & $(-0.7071,\,-0.7071,\,0)$\\
$\kappa_{\sigma}$ & 0.5 & $\hat{\rho}^{(3)}$ & $(0,\,1,\,0)$\\
$\kappa_{\mu}$ & 0.1 & &\\
	\hline \hline
\end{tabularx}
\end{table}

\newcolumntype{Y}{>{\raggedleft\arraybackslash}X}
\begin{table}[H]
	\centering
	\caption{Parameters for the clamp experiments of Section \ref{sec:results:clamp}.}
	\label{table:parameters_clamp}
	\vspace{-3.8pt}
	\begin{tabularx}{\columnwidth}{@{}l r | l r@{}}
		\hline \hline
		\textbf{Parameter} & \textbf{Value} & \textbf{Parameter} & \textbf{Value}\\
		\hline \hline
		$	N $& 120  & $\beta_{\mathrm{r},1}$ & 1\\
		$\Delta t$ & 0.033 & $\beta_{\mathrm{r},2}$ & 0.1 \\
		$n_{\mathrm{s}} $& 5  & $\beta_{\mathrm{r},3}$ & 100\\
		$	\vert \Pi \vert $& 6 & $\beta_{\mathrm{r},4}$ & 10\\
		
		$k_{\mathrm{t}}$ & 100&  $\beta_{\mathrm{c},1}$ & 1  \\
		$k_{\mathrm{r}}$ & 10 &$\beta_{\mathrm{c},2}$ & 0.1 \\
		$n_{\mathrm{hom}}$& 6 &$\beta_{\mathrm{c},3}$ & 10000 \\
		$\tau_{1} $& 0.0005 &$\beta_{\mathrm{c},4}$ & 1000 \\
		$\sigma_{1} $& \phantom{fillllllllllll}0.00025 &$\hat{\delta}$ & cf. Figure \ref{fig:realworldresults} \\
		$\mu_{\mathrm{init},1}$& 1 &  $\hat{\rho}^{(1)}$ & $(0,\,0,\,0)$  \\
		$\kappa_{\tau}$ & 0.5 & $\hat{\rho}^{(2)}$ & $(0.7071,\,0.7071,\,0)$\\
		$\kappa_{\sigma}$ & 0.5 & $\hat{\rho}^{(3)}$ & \phantom{$-$}$(-0.7071,\,0.7071,\,0)$\\
		$\kappa_{\mu}$ & 0.1 & $\hat{\rho}^{(4)}$ & $(0,\,1,\,0)$\\
		&  & $\hat{\rho}^{(5)}$ & $(0,\,-1,\,0)$\\
		\hline \hline
	\end{tabularx}
\end{table}

}

\bibliographystyle{ieeetr}
\bibliography{literature_ral25}

\newpage

\section{Biography Section}
\begin{IEEEbiography}[{\includegraphics[width=1in,height=1.25in,clip,keepaspectratio]{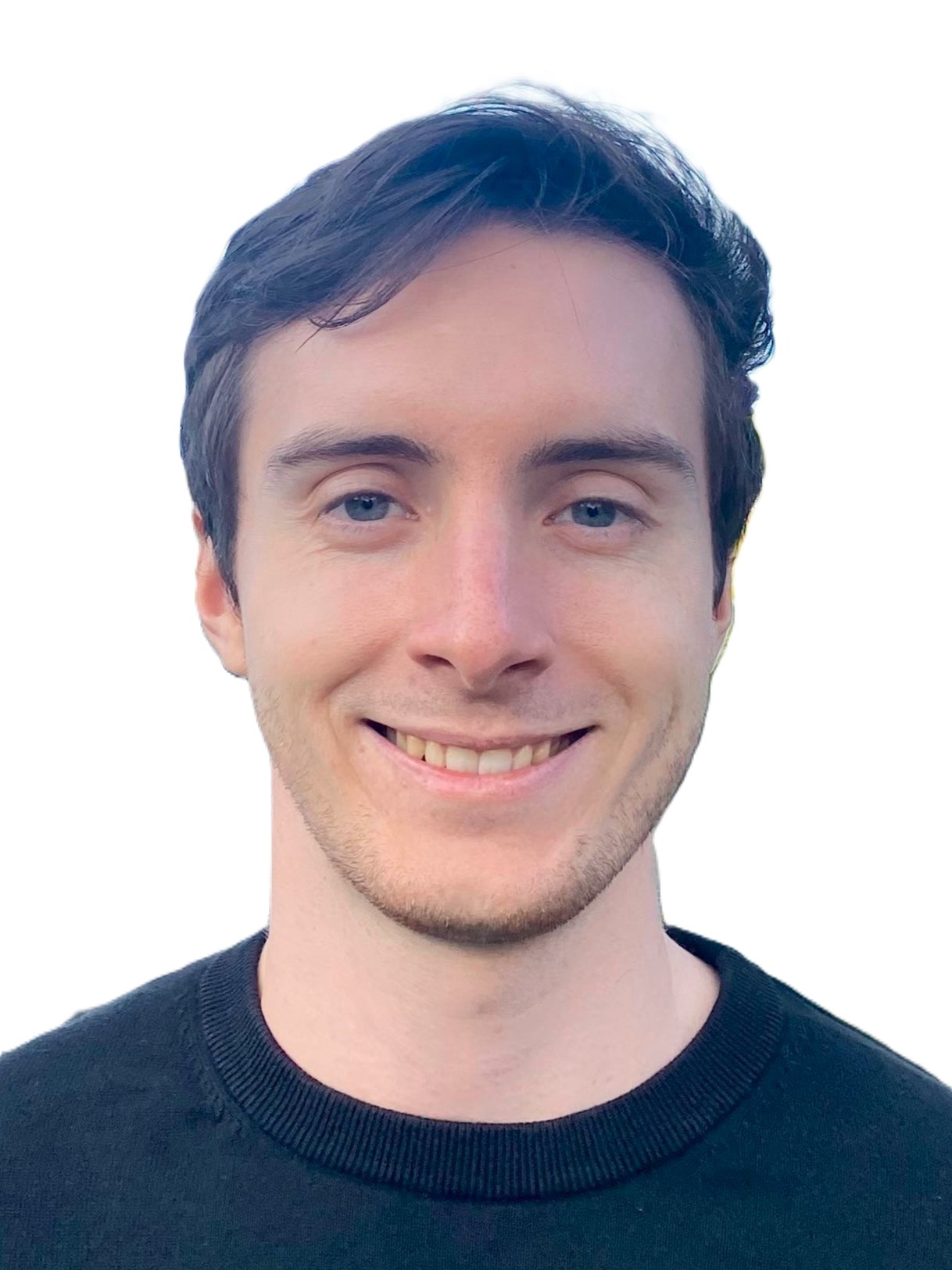}}]{Christian Dietz} received both a bachelor's and a master's degree in mathematics from the Technical University of Munich, Germany, in 2019 and 2022, respectively. Since 2022 he is an industrial PhD student associated with both Siemens and the Systems Control and Optimization Laboratory at the University of Freiburg, Germany. His research concentrates on numerical optimal control methods for contact-rich systems, with an application focus on industrial assembly planning.
\end{IEEEbiography}
\begin{IEEEbiography}[{\includegraphics[width=1in,height=1.25in,clip,keepaspectratio]{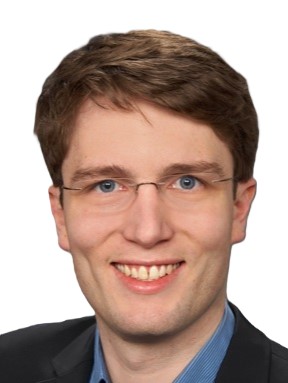}}]{Sebastian Albrecht} received the Ph.D. degree in Mathematics from Technische Universität München, Germany, in 2014. He joined Siemens AG, Munich, in 2015 as a Research Scientist, working in the areas of robotics, autonomous systems, and control. His research interests include numerical methods for nonlinear optimization and control, with a focus on their application to complex real‑world problems.
\end{IEEEbiography}
\begin{IEEEbiography}[{\includegraphics[width=1in,height=1.25in,clip,keepaspectratio]{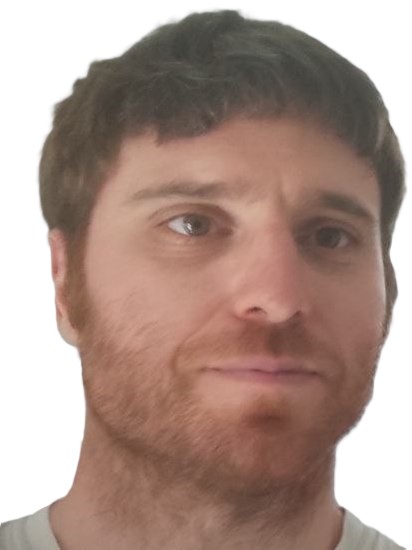}}]{Gianluca Frison}
	obtained a MSc in Automatic Control Engineering from the University of Padua (Italy), and a MSc in Mathematical Modeling and Computation from the Technical University of Denmark in 2012. In 2016 he obtained a PhD degree from Technical University of Denmark, with a doctoral thesis on algorithms and methods for high-performance model predictive control. He is currently employed as a post-doc at the University of Freiburg (Germany), working on high-performance linear algebra for embedded optimization with application to solvers for model predictive control, and as a senior developer at Mosek, working on the implementation of mathematical optimization algorithms.
\end{IEEEbiography}

\begin{IEEEbiography}[{\includegraphics[width=1in,height=1.25in,clip,keepaspectratio]{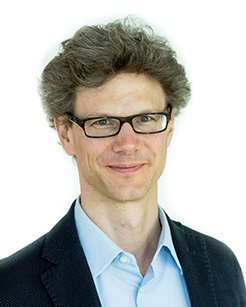}}]{Moritz Diehl} (Member, IEEE) received the dual
	Diploma degree in mathematics and physics from
	Heidelberg University, Heidelberg, Germany, and
	Cambridge University, Cambridge, U.K., in 1999,
	and the Ph.D. degree in optimization and nonlinear
	model predictive control from the Interdisciplinary
	Center for Scientific Computing, Heidelberg University, in 2001.
	From 2006 to 2013, he was a Professor with the
	Department of Electrical Engineering, KU Leuven
	University, Leuven, Belgium, where he was the
	Principal Investigator with the Optimization in Engineering Center (OPTEC).
	In 2013, he moved to the University of Freiburg, Freiburg, Germany, where
	he heads the Systems Control and Optimization Laboratory, Department of
	Microsystems Engineering (IMTEK), and is also with the Department of
	Mathematics. His research interests include optimization and control, spanning
	from numerical method development to applications in different branches of
	engineering, with a focus on embedded and renewable energy systems.
\end{IEEEbiography}
\begin{IEEEbiography}[{\includegraphics[width=1in,height=1.25in,clip,keepaspectratio]{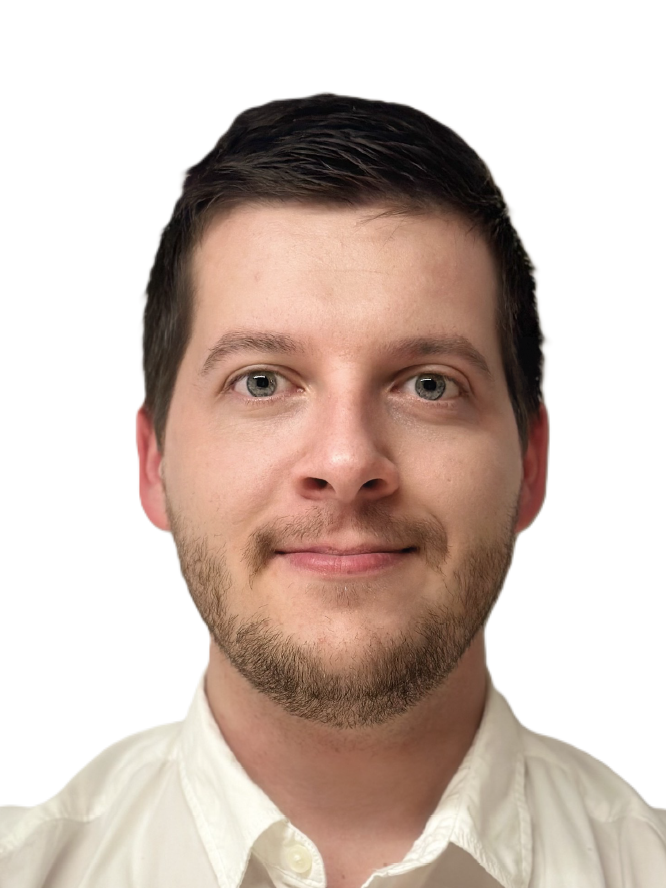}}]{Armin Nurkanovi\'c} received the B.Sc. degree from the Faculty of Electrical Engineering, Tuzla, Bosnia and Herzegovina, in 2015, and the M.Sc. degree in Electrical and Computer Engineering, Technical University of Munich, Munich, Germany, in 2018. In 2023, he received his Ph.D. degree in Engineering from the University of Freiburg, Freiburg, Germany. He received the IEEE Control Systems Letters Outstanding Paper Award in 2022 and was a finalist for the 2024 European Systems \& Control PhD Thesis Award. In 2025, he served as an interim professor of mathematical optimization at TU Braunschweig. His research interests include numerical methods for model predictive control, nonlinear optimization, and optimal control of nonsmooth and hybrid dynamical systems.
\end{IEEEbiography}

\vfill

\end{document}